\documentclass{article}

\PassOptionsToPackage{numbers, compress}{natbib}

\usepackage[preprint]{neurips_2025}

\usepackage[utf8]{inputenc} 
\usepackage[T1]{fontenc}    
\usepackage{hyperref}       
\usepackage{url}            
\usepackage{booktabs}       
\usepackage{amsfonts}       
\usepackage{nicefrac}       
\usepackage{microtype}      
\usepackage{graphicx}
\usepackage{adjustbox}
\usepackage{amsmath, amssymb, mathtools}
\usepackage{bm}
\usepackage{placeins}
\usepackage{tocloft}
\usepackage{fancyhdr}
\usepackage{etoc}
\usepackage{amsthm}
\usepackage{algorithm}
\usepackage{algpseudocode}
\usepackage{amsfonts}
\usepackage{times}
\usepackage{hhline}
\usepackage{multirow}
\usepackage{xcolor}         
\newcounter{codecounter} 
\newcommand{\codelink}[1]{%
    \stepcounter{codecounter}
    \textsuperscript{\thecodecounter}
    \fancyfoot[L]{\textsuperscript{\thecodecounter} Code is available at \url{#1}}
}
\title{Sparsification and Reconstruction from the Perspective of Representation Geometry}

\author{
  Wenjie Sun$^{1}$, Bingzhe Wu$^{2}$, Zhile Yang$^{1}$, Chengke Wu$^{1}$  \\
  $^{1}$Shenzhen Institute of Advanced Technology, CAS \\
  $^{2}$Shenzhen University \\
  \texttt{\{wj.sun, zl.yang, ck.wu\}@siat.ac.cn}, \texttt{wubingzheagent@gmail.com}
}

\begin{document}

\maketitle

\begin{abstract}
Sparse Autoencoders (SAEs) have emerged as a predominant tool in mechanistic interpretability, aiming to identify interpretable monosemantic features. However, how does sparse encoding organize the representations of activation vector from language models? What is the relationship between this organizational paradigm and feature disentanglement as well as reconstruction performance? To address these questions, we propose the SAEMA, which validates the stratified structure of the representation by observing the variability of the rank of the symmetric semipositive definite (SSPD) matrix corresponding to the modal tensor unfolded along the latent tensor with the level of noise added to the residual stream. To systematically investigate how sparse encoding alters representational structures, we define local and global representations, demonstrating that they amplify inter-feature distinctions by merging similar semantic features and introducing additional dimensionality. Furthermore, we intervene the global representation from an optimization perspective, proving a significant causal relationship between their separability and the reconstruction performance. This study explains the principles of sparsity from the perspective of representational geometry and demonstrates the impact of changes in representational structure on reconstruction performance. Particularly emphasizes the necessity of understanding representations and incorporating representational constraints, providing empirical references for developing new interpretable tools and improving SAEs. The code is available at \hyperlink{https://github.com/wenjie1835/SAERepGeo}{https://github.com/wenjie1835/SAERepGeo}.
\end{abstract}

\section{Introduction}
Mechanistic Interpretability (MI) aims to break down neural networks into interpretable components, analyzing their functions and interactions to understand the decision-making mechanisms of the entire network~\cite{37,38,39}. Sparse Autoencoders (SAEs), as a key tool in MI, the initial aim was to to address the polysemantic phenomenon caused by feature superposition~\cite{1}, where neural networks represent more independent features than neurons by assigning each feature a linear combination of neurons~\cite{23}. As a neural network implementation of dictionary learning, SAEs decompose input activation vectors into sparse and interpretable features, which often correspond to specific directions in the representation space~\cite{23}. To evaluate the effectiveness of this paradigm and provide optimization insights, we must understand how SAEs organize the representation of the activation vector.

A key claim of the linear representation hypothesis is that a model's state is a sparse linear combination of independent features~\cite{1,35,36}. Olah et al.~\cite{34} suggested that these features may lie on a manifold, where the angles between dictionary vectors of features are small, and adjacent features jointly respond to similar data. This led to the proposal of the feature manifold concept. Engels et al.~\cite{4} further validated this by performing clustering and dimensionality reduction of SAE dictionary elements, discovering circular-shaped, multidimensional features related to months and weekdays.

However, it is unclear whether the representation structure of concepts is formed during the language model's pre-training process or guided by the encoding mechanism of SAEs. Furthermore, how does the representation structure change during sparse encoding, and what is the relationship between these changes, disentangling polysemantic features, and reconstruction performance? Our work aims to explain the mechanisms of SAEs from the perspective of representation geometry and provide insights for their optimization.

\begin{enumerate}
    \item Inspired by Shrivastava et al.~\cite{9}, we represent the latent tensors of SAEs as points on a product manifold of symmetric semipositive definite (SSPD) matrices, where the rank variability of the SSPD matrices reflects the stratification of the manifold. Based on this, we propose the SAEManifoldAnalyzer (SAEMA) to analyze how pre-trained SAEs represent activation vectors corresponding to different concepts in language models. The results demonstrate the prevalence of stratified manifolds, where representations of the same concept are distributed across different manifolds.
    \item To understand how sparse encoding alters the representation structure, we define local and global representations, capturing the former by performing dimensionality reduction and clustering separately on the residual streams and the sparsely encoded latent tensors in the representation space. Specifically, sparse encoding reduces feature overlap by merging similar semantic features within local representations and introducing more dimensions into it. This finding reveals how SAEs achieve feature disentanglement to identify monosemantic features.
    \item To explore the relationship between the geometric properties of the representation structure and SAEs' reconstruction performance, we intervene in the geometric relationships among local representations from an optimization perspective. The results show that increased separability between different local representations causally enhances reconstruction performance. This finding underscores the necessity of incorporating geometric constraints into SAE optimization and provides new insights for mitigating the conflict between sparsity and reconstruction fidelity.
\end{enumerate}

\section{Representation Geometry Framework}

\subsection{Sparse Autoencoder}
Many recent studies has proposed employing Spaese Autoencoders (SAEs) to extract interpretable features from superposition for model understanding and editing, and the efficiency of this paradigm has been validated~\cite{40,41}. SAEs decompose each activation vector $\mathbf{x} \in \mathbb{R}^n$ into a sparse linear combination of learned feature directions using an overcomplete dictionary ($M \gg n$ basis vectors), transforming the complex activation space into sparse, interpretable representations through an encoder (Equation~\ref{eq:encoder})-decoder(Equation~\ref{eq:decoder}) architecture:

\begin{equation}
f(\mathbf{x}) := \sigma(\mathbf{W}_{\text{enc}} \mathbf{x} + \mathbf{b}_{\text{enc}}),
\label{eq:encoder}
\end{equation}
\begin{equation}
\hat{\mathbf{x}}(f(\mathbf{x})) := \mathbf{W}_{\text{dec}} f(\mathbf{x}) + \mathbf{b}_{\text{dec}},
\label{eq:decoder}
\end{equation}

where $\mathbf{x} \in \mathbb{R}^n$ denotes the language model's residual stream activation, $f(\mathbf{x}) \in \mathbb{R}^M$ represents the sparse feature activations, and $\sigma(\cdot)$ is a sparsity-enforcing nonlinearity. The encoder parameters $\mathbf{W}_{\text{enc}} \in \mathbb{R}^{M \times n}$, $\mathbf{b}_{\text{enc}} \in \mathbb{R}^M$ and decoder parameters $\mathbf{W}_{\text{dec}} \in \mathbb{R}^{n \times M}$, $\mathbf{b}_{\text{dec}} \in \mathbb{R}^n$ project between spaces of differing dimensionality. The columns of $\mathbf{W}_{\text{dec}}$ (denoted $\{\mathbf{d}_i\}_{i=1}^M$) form a feature dictionary used for reconstruction, with each basis vector $\mathbf{d}_i$ corresponding to a learned interpretable feature.

\subsection{SAE Manifold Analyzer (SAEMA)}
\begin{figure}
    \centering
    \includegraphics[width=1\linewidth]{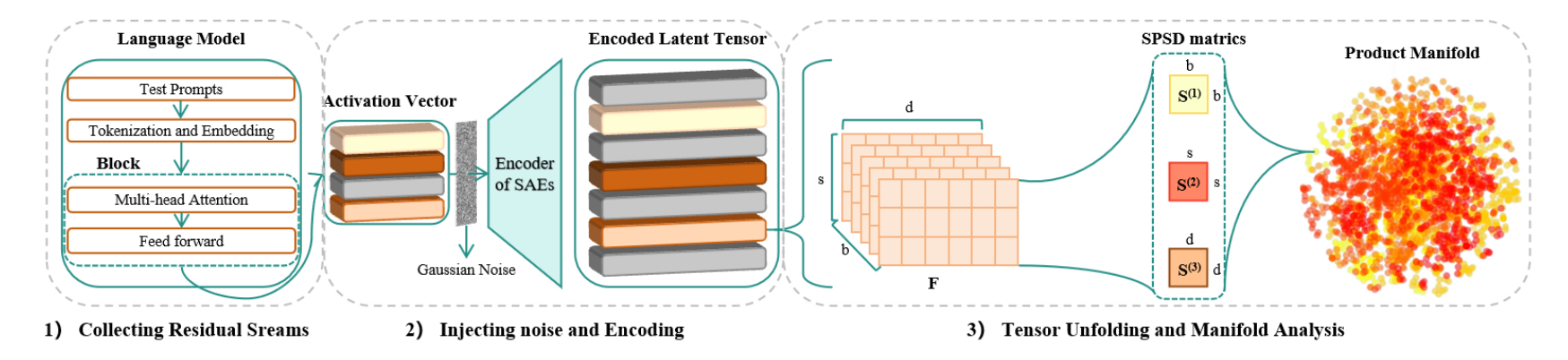}
    \caption{The workflow of SAEManifoldAnalyzer (SAEMA)}
    \label{fig1}
\end{figure}

SAEMA is a manifold learning-based method designed to analyze the manifold structure of concepts encoded by SAEs in language models. Figure~\ref{fig1} illustrates its workflow, which consists of the following three steps: $1)$ Collecting Residual Streams; $2)$ Injecting noise and Encoding; $3)$ Tensor Unfolding and Manifold Analysis.

\paragraph{Collecting Residual Streams}
For each model and concept, we extract activation vectors from the specified layer's residual stream hook (${hook\_resid\_post}$), generating a tensor with shape ${(batch\_size, seq\_len, d\_model)}$.

\paragraph{Injecting noise and encoding} Gaussian white noise is added to the residual stream in a feature-directed manner, with varying intensity levels. High-frequency features (the top 100 based on activation frequency) are perturbed with a noise intensity of \(2 \cdot \text{noise\_std}\), while the remaining features are subjected to a weaker noise level of \(0.2 \cdot \text{noise\_std}\). The perturbed residual stream is then processed by a pretrained Sparse Autoencoder (SAE), generating a latent tensor of shape \texttt{(batch\_size, seq\_len, d\_sae)}, where \(d_{\text{sae}}\) denotes the SAE’s feature dimension. To maintain computational efficiency, if \(d_{\text{sae}}\) exceeds 2048, the tensor is downsampled to 2048 features, selected based on their variance score in Equation~\ref{eq:downsampling}.

\begin{equation}\label{eq:downsampling}
\text{score}(j) = \text{Var}(f_j) \cdot \left(\sum_{i \neq j} |\text{Cov}(f_i, f_j)| + \epsilon \right)
\end{equation}

where $\text{Var}(f_j)$ is the variance of feature $j$, $\text{Cov}(f_i, f_j)$ is the covariance between features $i$ and $j$, and $\epsilon$ is a small constant to prevent division by zero.

\paragraph{Tensor Unfolding and Manifold Analysis} The latent tensor \(\mathcal{F}\) is unfolded along three modes (batch, sequence, feature) into matrices \(\{F^{(1)}, F^{(2)}, F^{(3)}\}\), and the corresponding SSPD matrices are computed:
    
\begin{equation}\label{eq:SSPD_compute}
S^{(i)} = \frac{F^{(i)}F^{(i)T} + (F^{(i)}F^{(i)T})^T}{2} + \epsilon I
\end{equation}
    
where \(\epsilon = 10^{-5}\) ensures numerical stability. Singular value decomposition is performed on each \(S^{(i)}\) to compute its rank:
    
\begin{equation}\label{eq:rank_SSPD}
\text{rank}(S^{(i)}) = \sum_{j=1}^{n} \mathbb{1}\left(\frac{\lambda_j}{\lambda_1} > \tau_i\right)
\end{equation}
    
where \(\lambda_j\) are the singular values (in descending order), and \(\tau_i\) is a dynamic threshold set as the first quartile of effective singular values (\(\lambda_j > 10^{-6} \lambda_1\)). The rank triplet \((r_1, r_2, r_3)\) characterizes the rank configuration of the latent tensor within the product manifold, reflecting its low-dimensional geometric structure and stability under noise perturbations. The detailed mathematical proof of how the variability of the rank of the SSPD matrix reflects the stratification of the product manifold can be found in the Appendix~\ref{proof_stratified manifold}.

\subsection{Local and Global Representation}
A key characteristic of hierarchical manifolds is being locally smooth yet globally non-smooth. To understand how sparse encoding reshapes representational structures comprehensively, we define both local and global representations.

\textbf{Definition 1}(Local Representation). The local representation is a low-dimensional subspace in the SAE latent space, corresponding to a semantically coherent feature cluster \(\mathcal{C}\) for a concept, identifiable through clustering, with each cluster described by an SSPD matrix from the feature-mode unfolding. For the latent tensor \(\mathcal{F}\)'s feature-mode \(F^{(3)}\), the \(k\)-th local representation of concept \(c\) is:
\[
\mathcal{M}_c^{(k)} = \left\{ \mathbf{z} \in \mathcal{C}_k \mid \operatorname{rank}(S^{(3)}(\mathbf{z})) = r_3^{(k)} \right\}.
\]

\textbf{Definition 2}(Global Representation). The global representation is the union of all local representations for a concept, described by a product manifold of SSPD matrices. For concept \(c\), the global representation is:
\[
\mathcal{G}_c = \bigcup_{k=1}^K \mathcal{M}_c^{(k)} \subset \prod_{r_3^{(k)}} S_{d_{\text{sae}}}^+(r_3^{(k)}).
\]

To quantify the geometric characteristics in local representations, we introduce the average intrinsic dimensionality (Avg. ID) to assess the effective dimensionality occupied by local representations, and Betti 0 to indicate the number of connected components within them. For global representations, we introduce the Minimum Spanning Tree Weight (MSTW), computed as the sum of Euclidean distances between the centers of different local representations, reflecting the overall dispersion of the global representation. Appendix~\ref{intro_eva_metx} provides the introduction of these metrics.

\section{Experiment Setup}

\subsection{Model and Dataset}\label{sec: model and dataset}
\paragraph{Model} We selected the post residual streams from the later layers of three language models for analysis: Layer 11 of GPT2-Small~\cite{44}, Layer 5 of Pythia-70M~\cite{45}, and Layer 19 of Gemma2-2B~\cite{46}, and loaded their corresponding pre-trained SAEs from SAE-Lens~\cite{7}. Table~\ref{tab: config of SAEs} and \ref{tab: config of language models} in Appendix~\ref{dataset and config} show the detailed configuration of SAEs and language models used in this work.
\paragraph{Dataset} We tested six concepts: Months, Weekdays, Chemical Elements, Color, Planets, Number, Alphabet, Phonetic Symbol and constellations. The selection criteria were based on whether their keywords formed clusters in the representation space. For these concepts, we designed diverse and information-dense prompts (Table~\ref{tab:prompts_wrapped} in Appendix~\ref{dataset and config} shows the prompts of Months, Weekdays and Chemical Elements) to capture as complete a representation as possible. The final input to the SAEs was the masked residual stream, filtered of irrelevant tokens (such as prepositions, conjunctions, and <|endoftext|> et al).

\subsection{Experimental Design}\label{sec: exp design}
In this section, we present three experimental designs to progressively verify our theoretical contributions. Specifically, in Section~\ref{case1_design}, we use SAEMA to test the manifold patterns of different concepts across various language models to verify the universality of stratified manifolds. In Section~\ref{case2_design}, we perform dimensionality reduction and density-based clustering on both residual stream and latent tensors in the representation space to reveal how sparse encoding modifies the geometric structures of both local and global representations. To investigate the relationship between representation geometry and reconstruction performance, in Section~\ref{case3_design}, we perform intervention to the global representations from an optimization perspective, aiming to prove the causal relationship between separability and reconstruction performance.

\subsubsection{Case 1: Stratified Manifold}\label{case1_design}
To understand how sparse encoding organizes the representations of activation vectors, we used the SAEMA module to test the representation of all concepts by three SAEs mentioned in Section~\ref{sec: model and dataset} and pre-trained on residual streams of different language models—under varying levels of noise. White Gaussian noise perturbations with nine different standard deviations, ranging from 0.0 to 10.0 (0.0, 0.02, 0.05, 0.1, 0.2, 0.5, 1.0, 2.0, 5.0, 10.0), were added to each of the nine model-concept pairs. To prevent the sparsity constraint of the SAEs from filtering low-frequency activations, we specifically applied 2x noise to high-frequency activations. By observing the variability of the ranks $(r_{1}, r_{2}, r_{3})$ of the SSPD matrices corresponding to different modal tensors with respect to noise, to verify whether the latent tensors span manifold layers of different rank configurations.

\paragraph{Evaluation metrics}\

Rank Variability: The variability of the rank triplet (r$_1$, r$_2$, r$_3$) is examined, with a focus on r$_3$ (feature mode), to verify if latent representations traverse different manifold strata.

Average Geodesic Distance (AGD): To quantify the continuous geometric differences between SSPD matrices on the manifold and provide supplementary evidence for rank variability, AGD is defined in this paper as the mean geodesic distance of SSPD matrices on the $S^{+}_{n}$ manifold. The specific formula can be found in the Appendix~\ref{intro_eva_metx}.

\subsubsection{Case 2: Changes in Representation Structure}\label{case2_design}

To investigate the impact of sparse encoding on the structure of representations, we compare representations before (residual stream) and after (SAE latent tensor) sparse encoding. Selecting the same models and three concepts (Months, Weekdays and Chemical Elements) in Case 1. Aiming to reveal the way that sparse encoding organizes representations. The specific experimental steps are as follows:

\textbf{Dimensionality Reduction:} Residual representations $\mathbf{x} \in \mathbb{R}^{{batch\_size} \times {seq\_len} \times {d\_model}}$ and SAE latent tensors $\mathcal{F} \in \mathbb{R}^{{batch\_size} \times {seq\_len} \times {d\_sae}}$ are extracted from the zero-noise cache of Case 1. Representations before and after sparse encoding are first flattened to $\mathbb{R}^{{batch\_size} \times {seq\_len} \times {d\_sae}}$, normalized, and then reduced to 50 dimensions with UMAP, balancing computational efficiency while preserving the structure of the representations as much as possible.

\textbf{Clustering:} HDBSCAN (minimum cluster size = 10) is applied to the dimensionality-reduced representations to identify individual local representations (clusters). We then estimate the intrinsic dimensionality of each local representation and computed its persistent homology to obtain the corresponding Betti 0 (connected components), thereby evaluating the structural changes induced by sparse encoding on local representations. Furthermore, to quantify changes in local representations, we compute the MSTW for both the residual stream and latent tensors, and use Procrustes analysis to compare the geometric rearrangement of the centers of the two sets of local representations before and after sparse encoding. Regarding the formula and definition of Procrustes Disparity can be found in Appendix~\ref{intro_eva_metx}.

\subsubsection{Case 3: Intervention Representation Structure}\label{case3_design}
To explore the relationship between the geometric structure of latent representations in SAEs and their reconstruction performance, we intervene in the representation structure from an optimization perspective. Inspired by Lee et al.~\cite{42} and Nao et al.~\cite{43}, we employ the Gromov-Wasserstein distance ($d_{GW}$) to compare the distance metrics of the metric spaces before and after the intervention and incorporate it as an optimization term in the loss function, aim to explore the geometric properties between local representations and reconstruction performance while preserving the representation structure as much as possible. The detailed optimization strategies are as follows:

\paragraph{Procedure:} We compute the centers of various local representations of concepts after sparse encoding based on the HDBSCAN cluster labels from Case 2:
\begin{equation}\label{eq:center_compute}
\mathbf{c}_i = \frac{1}{|\mathcal{C}_i|} \sum_{\mathbf{z} \in \mathcal{C}_i} \mathbf{z}, 
\end{equation}

where \(\mathcal{C}_i\) denotes the set of latent representations in cluster $i$, and \(\mathbf{z} \in \mathbb{R}^{d_{\text{sae}}}\) is the flattened latent representation with shape \((\text{batch\_size} \cdot \text{seq\_len}, d_{\text{sae}})\), representing the arithmetic mean of the local semantics of the given concept. The intervention is achieved by translating these centers and their corresponding local representations by different step size $\alpha$, and the translation direction of was guided by applying stochastic gradients ($grad$) to \(\mathbf{c}_i\) to minimize \(L\) in Equation.\ref{eq:gw_loss} during optimization, expressed as $c'_{i}\gets c_{i}+\alpha \times grad$.
\begin{equation}
L = d_{\text{GW}}(D_{\text{original}}, D_{\text{intervened}}) + \lambda_{MSE} \text{MSE}(\mathbf{x}, \hat{\mathbf{x}})
\label{eq:gw_loss}
\end{equation}

\begin{equation}
d_{\text{GW}}(D_{\text{original}}, D_{\text{intervened}}) = \min_{\pi \in \Pi(\mu, \nu)} \sum_{i,j,i',j'} |D_{\text{original}}(i,j) - D_{\text{intervened}}(i',j')| \pi_{i,i'} \pi_{j,j'}
\label{eq:GWD}
\end{equation}

\begin{equation}
D_{\text{original}}(i,j) = \frac{\|\mathbf{z}_i - \mathbf{z}_j\|_2}{\max_{i,j} \|\mathbf{z}_i - \mathbf{z}_j\|_2}, D_{\text{intervened}}(i',j') = \frac{\|\mathbf{z}_i' - \mathbf{z}_j'\|_2}{\max_{i',j'} \|\mathbf{z}_i' - \mathbf{z}_j'\|_2}
\label{eq: EDM}
\end{equation}

where \(d_{\text{GW}}\) in Equation~\ref{eq:GWD} is the GW distance, measuring the structural dissimilarity between the original and intervened latent representations. \(D_{\text{original}}\) and \(D_{\text{intervened}}\) are the normalized Euclidean distance matrices of the original and intervened latent representations, defined in Equation~\ref{eq: EDM}. $\lambda_{MSE}$ is used to control the weight of Mean Squared Error (MSE), set to 1.

\paragraph{Evaluation Metrics:} During the optimization process, local representations are translated along with their $\mathbf{c}_i$, preserving the intra-cluster aggregation degree. To quantify the separability between different local representations, we compute the Average Euclidean Distance between Pairs (AEDP) of cluster centers. The reconstruction performance was quantified by MSE. The detailed introduction about AEDP and MSE can be found in Appendix~\ref{intro_eva_metx}.

\section{Experimental Results and Analysis}

\subsection{Case 1}\label{mainsec: case1}
\begin{figure}[h]
    \centering
    \includegraphics[width=1.0\linewidth]{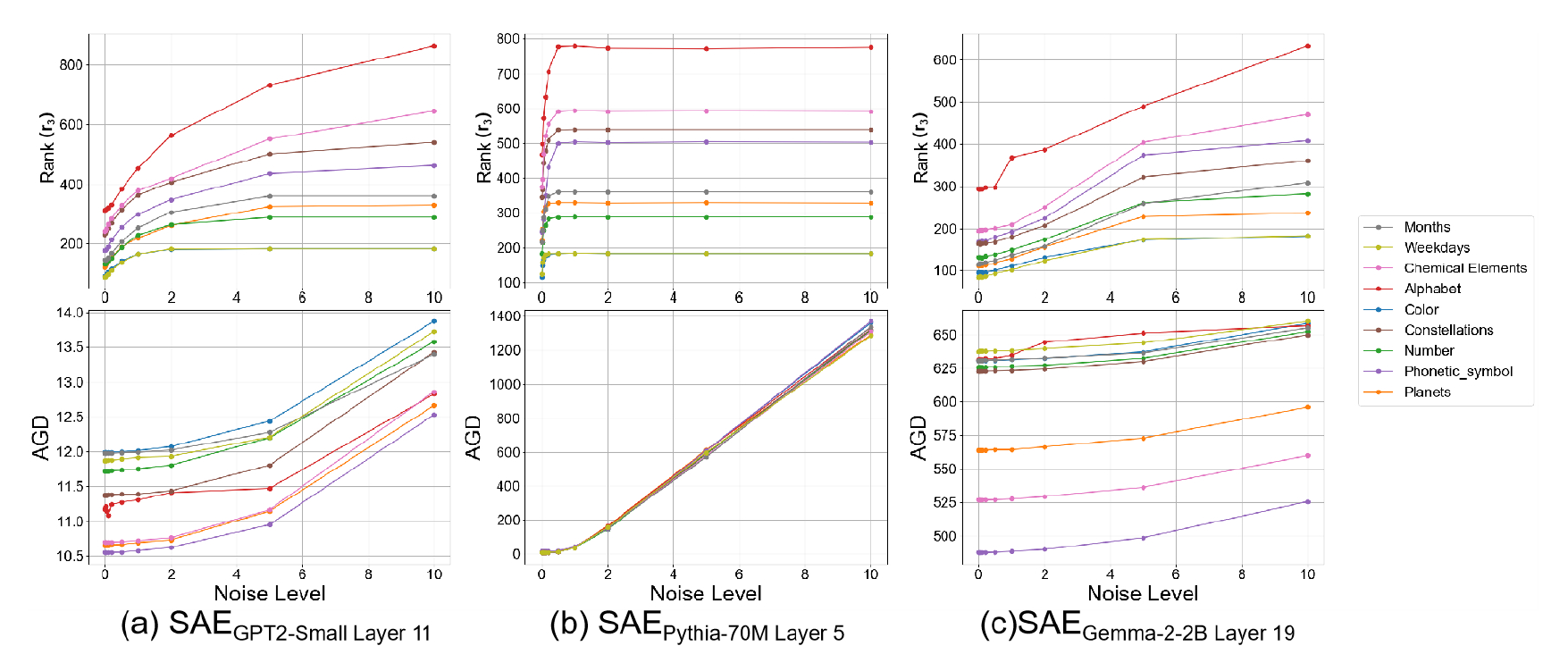}
    \caption{The changes of $r_{3}$ and AGD corresponding to different concepts encoded by pre-trained SAEs under different noise levels.}
    \label{fig:noise_vs_rank_and_AGD}
\end{figure}

Figure~\ref{fig:noise_vs_rank_and_AGD} illustrates the variability of $r_{3}$ (the rank of SSPD matrix of feature modal) and AGD for three SAEs representing the concepts under different noise levels. Detailed experimental results can be found in Appendix~\ref{exp:case1}. We observe that the \texttt{batch\_size} and \texttt{seq\_len} modal ranks ($r_{1}$, $r_{2}$) remain constant, which may be attributed to the fact that prompts with high information density and strong semantic consistency stabilize the covariance structure. In contrast, the feature modal rank ($r_{3}$) exhibits varying degrees of variability ($SAE_{Pythia-70M\ Layer\ 5}$: 57.37\%, $SAE_{GPT2-Small\ Layer\ 11}$: 140.72\%,  $SAE_{Gemma-2-2B\ Layer\ 19}$: 141.67\%), proving that the representational structure of SAEs is stratified.

Additionally, the initial AGD values under zero noise reflect the compactness of the manifolds, showing that larger models tend to have more dispersed representations. The variability of AGD also demonstrates a negative correlation with model scale, the larger the language model, the smaller the AGD variability (Pythia-70M: 145x increase, GPT2-Small: 15.85\% increase, Gemma-2-2B: 4.53\% increase). This may be attributed to the more dispersed representations mitigating the impact of noise. Regarding the sharp AGD surge in Pythia-70M, may be due to high noise causing geometric collapse in the manifold.

\subsection{Case 2}\label{mainsec: case2}

\begin{table}[h]
\centering
\caption{Changes in Local and Global Representation Structures Before and After sparse encoding}
\label{tab:exp-case2}
\begin{adjustbox}{width=\textwidth}
\begin{tabular}{ccccccc}
\toprule
\multirow{2}{*}{\textbf{Model}} & \multirow{2}{*}{\textbf{Concept}} & \multirow{2}{*}{\textbf{Clusters}} & \multicolumn{2}{c}{\textbf{Local Representation}} & \multicolumn{2}{c}{\textbf{Global Representation}} \\
\cmidrule(lr){4-5} \cmidrule(lr){6-7}
& & & \textbf{Avg. ID$\uparrow$} & \textbf{Betti 0$\uparrow$} & \textbf{MSTW$\downarrow$} & \textbf{Procrustes Disparity} \\
\midrule
\multirow{3}{*}{GPT2-Small Layer 11} & Months & (22, 18) & (3.01, 3.25) & (21.09, 26.11) & (196.57, 101.16) & 0.61 \\
& Weekdays & (9, 9) & (3.29, 3.78) & (25.67, 26.00) & (94.57, 73.61) & 0.30 \\
& Chemical Elements & (27, 23) & (2.58, 2.86) & (30.74, 36.74) & (119.87, 130.50) & 0.63 \\
\midrule
\multirow{3}{*}{Pythia-70M Layer 5} & Months & (18, 17) & (3.01, 3.15) & (26.67, 28.24) & (165.13, 90.95) & 0.52 \\
& Weekdays & (6, 8) & (3.32, 3.81) & (35.83, 27.38) & (59.32, 55.21) & 0.15 \\
& Chemical Elements & (22, 26) & (3.19, 2.62) & (34.55, 27.19) & (96.80, 74.80) & 0.61 \\
\midrule
\multirow{3}{*}{Gemma-2-2B Layer 19} & Months & (21, 18) & (2.77, 3.00) & (18.81, 22.00) & (199.27, 113.75) & 0.40 \\
& Weekdays & (8, 6) & (3.80, 4.24) & (28.88, 39.67) & (80.46, 67.67) & 0.18 \\
& Chemical Elements & (23, 22) & (2.89, 3.13) & (26.45, 27.43) & (172.65, 102.28) & 0.53 \\
\bottomrule
\end{tabular}
\end{adjustbox}
\end{table}

The cluster counts in Table~\ref{tab:exp-case2} generally exhibit variability after sparse encoding, with 7 out of 9 model-concept pairs showing a decrease or constant in the number of clusters. Interestingly, the number of clusters also shows a certain positive correlation with the semantic complexity of the concepts (Chemical Elements > Months > Weekdays). For example, concept of chemical element that contains the most keywords corresponds to the largest number of clusters.

\textbf{Local Representation:} Seven out of nine model-concept pairs demonstrate an increase in Avg. ID, indicating that sparse encoding introduces more degrees of freedom into local representations, possibly due to the overcomplete basis vector of SAEs introducing additional feature dimensions. Simultaneously, seven pairs show an increase in Betti 0, i.e., the number of connected components, suggesting that local representations are segmented into more substructures. The increase in Avg. ID and connected components indicates that substructures within the same local representation become more numerous and the differences between substructures become greater, which means the overlap between features is reduced. This phenomenon indirectly explain the mechanism of feature disentanglement. Detailed discussion can be seen in the Appendix~\ref{app: discussion_case2}. 

\textbf{Global Representation:} Since sparse encoding alters the number of local representations, Procrustes analysis was performed considering only the labels of valid clusters. The results demonstrate the universality of topological rearrangement. MSTW serve as a metric to quantify the dispersion of global representations, showed a decreasing trend in 8 out of 9 model-concept pairs, indicating that the global representational structure was compressed during this process. The increase in the Avg. ID of local representations, alongside the compression of global representations, appears contradictory at first glance. However, this suggests that sparse encoding achieves disentanglement solely through effective dimensions, i.e., most dimensions are wasted. It also implies that even with overcomplete basis vector, this approximation method still cannot avoid the representation compression phenomenon of the autoencoder architecture.

\subsection{Case 3}\label{sec:main_case3_discussion}

Table~\ref{tab:case3_exp_resu} shows the variations of $d_{GW}$, MSE, and AEDP for different values of $\alpha$. All nine model-concept pairs exhibit a trend where AEDP increases and MSE decreases as $\alpha$ increases. This indicates that during the optimization process, the translation directions of each local representation tend to move away from each other, thereby increasing separability. Figure~\ref{fig:case3_dGW_AEDP_vs_Alpha} in Appendix~\ref{exp:case3} shows how AEDP and $d_{GW}$ change as $\alpha$ increases. While increasing $\alpha$ led to an increase in $d_{GW}$ for $SAE_{ GPT2-Small\ Layer\ 11}$ and $SAE_{Gemma-2-2B\ Layer\ 19}$, this was not the case for $SAE_{Pythia-70M\ Layer\ 5}$. Interestingly, despite the inconsistent trends of $d_{GW}$ with respect to $\alpha$ across the three models, they consistently show a positive correlation between AEDP and $\alpha$. This implies that an increase in $\alpha$ does not necessarily lead to an increase in the geometric structure of the representations but directly increases the separability between local representations. Furthermore, Figure~\ref{fig:case3_MSE_vs_dGW_AEDP} in Appendix~\ref{exp:case3} shows the variations in MSE with respect to AEDP and $d_{GW}$ for different $\alpha$ values. The changes in MSE and $d_{GW}$ do not show a consistent pattern across the three models, but they all consistently exhibit a trend where MSE decreases as AEDP increases. Specifically, the Pearson correlation coefficient between AEDP and MSE is -0.89 for $SAE_{ GPT2-Small\ Layer\ 11}$, -0.97 for $SAE_{Pythia-70M\ Layer\ 5}$, and -0.93 for $SAE_{Gemma-2-2B\ Layer\ 19}$, all showing significant negative correlations.

To validate the causal relationship between AEDP and MSE, we modified the loss function in Equation~\ref{eq:gw_loss} by replacing \(d_{GW}\) with \(AEDP^{-1}\) while keeping the optimization approach unchanged, as detailed in Appendix~\ref{case3.2}. The results indicate that the contribution of the \(AEDP^{-1}\) term to the total loss is almost negligible. However, MSE and AEDP still exhibit a significant negative correlation. In other words, during optimization, MSE dominates the total loss function to steer local representations away from each other, thereby achieving its own reduction. Based on this, we conclude that the increase in the separability of local representations causally enhances reconstruction performance.

\begin{table}[h]
\centering
\caption{Variations of MSE and AEDP corresponding to different translation step sizes ($\alpha$) during optimization}
\label{tab:case3_exp_resu}
\begin{adjustbox}{width=\textwidth}
\begin{tabular}{lccccccccccccc}
\toprule
\multirow{2}{*}{Conceptual Category} & \multirow{2}{*}{$\alpha$} & \multicolumn{4}{c}{$SAE_{GPT2-Small\ Layer\ 11}$} & \multicolumn{4}{c}{$SAE_{Pythia-70M\ Layer\ 5}$} & \multicolumn{4}{c}{$SAE_{Gemma-2-2B\ Layer\ 19}$} \\ \cmidrule(lr){3-6} \cmidrule(lr){7-10} \cmidrule(lr){11-14}
& & $d_{GW}$ & MSE$\downarrow$ & Orig. AEDP & AEDP$\uparrow$ & $d_{GW}$ & MSE$\downarrow$ & Orig. AEDP & AEDP$\uparrow$ & $d_{GW}$ & MSE$\downarrow$ & Orig. AEDP & AEDP$\uparrow$ \\ \midrule
\multirow{5}{*}{Months} & 0.5 & 0.004 & 58.09 & 10.50 & 11.32 & 0.071 & 470.77 & 7.19 & 8.21 & 0.372 & 44.28 & 358.97 & 503.41 \\
& 0.8 & 0.015 & 55.82 & 10.50 & 12.42 & 0.065 & 385.93 & 7.19 & 9.55 & 0.401 & 42.44 & 358.97 & 636.98 \\
& 1.0 & 0.027 & 53.32 & 10.50 & 13.33 & 0.045 & 329.73 & 7.19 & 10.63 & 0.417 & 42.20 & 358.97 & 731.61 \\
& 1.2 & 0.038 & 52.61 & 10.50 & 14.32 & 0.031 & 299.82 & 7.19 & 11.79 & 0.413 & 40.13 & 358.97 & 826.48 \\
& 1.5 & 0.064 & 48.68 & 10.50 & 16.01 & 0.059 & 247.75 & 7.19 & 13.68 & 0.420 & 40.09 & 358.97 & 838.85 \\
\midrule
\multirow{5}{*}{Weekdays} & 0.5 & 0.002 & 56.04 & 11.58 & 12.38 & 0.073 & 536.59 & 7.46 & 8.48 & 0.243 & 41.36 & 774.53 & 888.30 \\
& 0.8 & 0.007 & 54.19 & 11.58 & 13.25 & 0.085 & 460.20 & 7.46 & 9.79 & 0.258 & 39.24 & 774.53 & 978.66 \\
& 1.0 & 0.014 & 52.69 & 11.58 & 14.29 & 0.110 & 440.71 & 7.46 & 10.84 & 0.262 & 36.56 & 774.53 & 1071.95 \\
& 1.2 & 0.030 & 51.97 & 11.58 & 15.29 & 0.057 & 386.57 & 7.46 & 12.02 & 0.264 & 36.09 & 774.53 & 1124.52 \\
& 1.5 & 0.052 & 48.34 & 11.58 & 16.91 & 0.095 & 331.07 & 7.46 & 13.89 & 0.267 & 37.96 & 774.53 & 1241.20 \\
\midrule
\multirow{5}{*}{Chemical Elements} & 0.5 & 0.010 & 51.01 & 8.49 & 9.53 & 0.017 & 338.21 & 7.53 & 8.58 & 0.310 & 41.23 & 318.12 & 467.12 \\
& 0.8 & 0.027 & 50.22 & 8.49 & 10.81 & 0.016 & 309.30 & 7.53 & 9.92 & 0.367 & 40.57 & 318.12 & 604.61 \\
& 1.0 & 0.049 & 49.23 & 8.49 & 11.82 & 0.011 & 262.40 & 7.53 & 10.99 & 0.358 & 40.34 & 318.12 & 612.10 \\
& 1.2 & 0.077 & 48.53 & 8.49 & 12.94 & 0.008 & 253.90 & 7.53 & 12.15 & 0.370 & 40.12 & 318.12 & 727.67 \\
& 1.5 & 0.109 & 48.02 & 8.49 & 14.78 & 0.044 & 244.54 & 7.53 & 14.03 & 0.369 & 40.23 & 318.12 & 699.76 \\
\bottomrule
\end{tabular}
\end{adjustbox}
\end{table}

\section{Related Work}
\subsection{Sparse Autoencoder}
SAEs, as a critical tool in mechanistic interpretability, aim to address the polysemanticity problem in LLMs caused by feature superposition. However, this paradigm faces several challenges, including shrinkage bias induced by L1 regularization, the trade-off between sparsity and reconstruction fidelity, and the presence of dead features during training. Specifically, shrinkage bias refers to the systematic underestimation of feature activation values by L1 regularization, which degrades reconstruction quality. Rajamanoharan et al.~\cite{27} proposed Gated SAE, introducing a gating function and a magnitude estimation function to mitigate this issue. Additionally, the balance between sparsity and reconstruction fidelity limits the scalability of SAEs. To address this, Taggart et al.~\cite{31} introduced ProLU SAE, replacing the ReLU activation function with a ProLU activation function that learns dynamic thresholds. Similarly, Rajamanoharan et al.~\cite{8} employed the JumpReLU activation function, which sets a fixed activation threshold and uses L0 regularization to control and avoid biases introduced by small activation values. To tackle the inefficiency of dead features and the forced fixed number of activated features for all inputs, Bussmann et al.~\cite{28} proposed BatchTopK SAE, which applies TopK selection across the entire batch, allowing different samples to activate varying numbers of features. Ayonrinde et al.~\cite{32} ensured feature utilization efficiency by constraining the proportion of samples each feature activates. Significant research has also been devoted to addressing cross-layer challenges in mechanistic interpretability, such as Layer Group SAE~\cite{29}, which trains shared SAEs by computing activation similarities between layers, and Switch SAE~\cite{30}, which draws inspiration from the Mixture of Experts framework.

\subsection{Representation Geometry}
The relationship between representation in latent space and task performance has been well-established in neuroscience\cite{33} and statistical mechanics. Numerous studies have explored language model principles through the lens of representational geometry. For instance, Ethayarajh et al.~\cite{6} discovered that contextual representations exhibit anisotropy across all non-input layers and capture semantic information more effectively than static word embeddings. Chang et al.~\cite{5} identified stable language-neutral axes in linguistic subspaces across different languages, while Shai et al.~\cite{3} demonstrated the linear existence of Belief State Geometry in Transformer residual streams. SAEs trained on language model activation vectors have also revealed intriguing representational structures. Engles et al.~\cite{4}, through clustering and dimensionality reduction of dictionary elements, uncovered circular features related to temporal concepts such as days and months. Li et al.~\cite{2} identified shared directions for opposing concepts, forming parallelogram structures. Bricken et al.~\cite{23} revealed compressed global structures by applying 2D UMAP transformations to SAE encoding directions. However, the geometric transformations during sparse encoding and reconstruction, as well as their relationship to disentanglement, remain insufficiently explored. Given the current substantial research investment in SAEs, we believe it is necessary to provide an optimized reference from the perspective of representation geometry.

\subsection{Disentangled Representation Learning}
Disentangled representation learning, as a crucial component of unsupervised learning, aims to separate latent factors in data into independent and interpretable representations, thereby enhancing model interpretability~\cite{12, 13}. This concept has driven research in natural language processing to disentangle semantic and contextual elements, offering new perspectives on the interpretability and editability of language models. Mairal et al.~\cite{15} first introduced the dictionary learning to natural language processing, utilizing overcomplete basis vectors to decompose inputs into sparse linear combinations, generating sparse representations of textual features and laying the foundation for subsequent semantic decomposition.

Regularization serves as a key mechanism for achieving disentanglement~\cite{16}, prompting researchers to explore additional regularization architectures such as variational autoencoders~\cite{17,18,19} and generative adversarial networks~\cite{20,21,22}. In recent years, SAEs have garnered significant attention due to their sparsity and interpretability in disentangling language model representations. Cunningham et al.~\cite{23} pioneered the application of SAEs to LLMs, decomposing activations into sparse, monosemantic features, thereby enhancing interpretability in the era of large language models. With the successful practice of scaling laws, the approximately overcomplete basis vectors in SAEs introduce a conflict between sparsity and reconstruction performance \cite{24}. Moreover, as an unsupervised learning method, the evaluation of feature monosemanticity presents a bottleneck for further replication~\cite{25,26}.

\section{Conclusion}
Our work addresses three key questions sequentially: how sparse coding structures representations, how SAEs achieve feature disentanglement, and the relationship between representational geometry and SAEs reconstruction performance. We observed that SAEs exhibit stratified representations for different concepts within language models, meaning that the representations of the same concept are dispersed across different manifolds. Furthermore, by applying dimensionality reduction and clustering to activation vectors before and after sparse encoding to capture local representations, we find that sparse encoding reduces feature overlap by merging similar semantic features within local representations and introducing more dimentionality, thereby achieving feature disentanglement. Finally, through an optimization-based intervention on the geometric relationships among local representations within the global representation, we significantly improved the reconstruction performance of SAEs without any post-training, demonstrating that increased separability between local representations causally leads to a decrease in reconstruction performance.

In summary, our findings offer guidance for improving SAEs and developing novel tools for decomposing neural networks from the perspective of representation geometry, particularly emphasizing the necessity of understanding and incorporating representation geometry constraints.

\newpage
\bibliographystyle{IEEEtran}  
\small
\bibliography{Reference}

\normalsize


\appendix

\clearpage

\section*{Appendix}
\addcontentsline{toc}{section}{Appendix}
\renewcommand{\cftsecleader}{\cftdotfill{\cftdotsep}}
\setlength{\cftbeforesecskip}{1pt}
\setlength{\cftaftertoctitleskip}{1pt}
{\small 
\begin{center}
\textbf{Appendix Contents}
\end{center}
\begin{itemize}
  \item[A] Proof: The variability of $Rank(S^{(i)}$ reflects product manifold stratification \dotfill \pageref{proof_stratified manifold}
  \item[B] Experimental Details and Results \dotfill \pageref{app:Experimental Details and Results}
  \begin{itemize}
    \item[B.1] Datasets and Configurations \dotfill \pageref{dataset and config}
    \item[B.2] Experiment Results of Case 1 \dotfill \pageref{exp:case1}
    \item[B.3] Discussion of the results of the experiment in Case 2 \dotfill \pageref{app: discussion_case2}
    \item[B.4] Experiment Results of Case 3 \dotfill \pageref{app:exp_resu_case3}
  \end{itemize}
  \item[C] Evaluation Metrics \dotfill \pageref{intro_eva_metx}
  \item[D] Limitations \dotfill \pageref{limitation}
  \item[E] Impact Statement \dotfill \pageref{broader impact}
  \item[F] Reproducibility \dotfill \pageref{reproducibility}
\end{itemize}
\vspace{1pt}
}

\section{Proof: The variability of $Rank(S^{(i)}$) reflects product manifold stratification}\label{proof_stratified manifold}

For latent tensor \( \mathcal{F} \in \mathbb{R}^{I_1 \times I_2 \times I_3} \), where \( I_1 = {batch\_size} \), \( I_2 = {seq\_len} \), and \( I_3 = d_{{sae}} \). The mode-\( i \) unfolding \( F^{(i)} \) is a matrix obtained by arranging the fibers of \( \mathcal{F} \) along mode \( i \). For example, \( F^{(3)} \in \mathbb{R}^{I_3 \times (I_1 I_2)} \) stacks all feature vectors across batch and sequence dimensions.

The SSPD matrix is defined as:
\[
S^{(i)} = \frac{F^{(i)} F^{(i)^T} + (F^{(i)} F^{(i)^T})^T}{2} + \epsilon I = F^{(i)} F^{(i)^T} + \epsilon I,
\]
where \( \epsilon = 10^{-5} \) ensures numerical stability, and the symmetry follows since \( (F^{(i)} F^{(i)^T})^T = F^{(i)} F^{(i)^T} \).

Since \( \epsilon I \) is a small perturbation, the rank of \( S^{(i)} \) approximates the rank of \( F^{(i)} F^{(i)^T} \). By properties of matrix products, \( \text{rank}(F^{(i)} F^{(i)^T}) = \text{rank}(F^{(i)}) \), because \( F^{(i)^T} \) maps the row space of \( F^{(i)} \) to itself. Thus:
\[
r_i = \text{rank}(S^{(i)}) \approx \text{rank}(F^{(i)}).
\]

The \( \text{rank}(F^{(i)}) \) is the dimension of the column space of \( F^{(i)} \), which represents the subspace spanned by the mode-\( i \) fibers of \( \mathcal{F} \). For mode 3 (features), \( F^{(3)} \) contains feature vectors across all tokens, and \( \text{rank}(F^{(3)}) \) is the number of linearly independent feature directions, reflecting the effective dimensionality of the latent feature space. Thus, \( r_i \) characterizes the effective dimension of the mode-\( i \) subspace of \( \mathcal{F} \), and the triplet \( (r_1, r_2, r_3) \) defines the geometric structure of \( \mathcal{F} \) on the product manifold \( S_{I_1}(r_1) \times S_{I_2}(r_2) \times S_{I_3}(r_3) \).

For the latent tensor \( \mathcal{F} \in \mathbb{R}^{I_1 \times I_2 \times I_3} \), with mode-\( i \) unfolding \( F^{(i)} \), and SSPD matrix \( S^{(i)} = F^{(i)} F^{(i)^T} + \epsilon I \). The rank \( r_i = \text{rank}(S^{(i)}) \approx \text{rank}(F^{(i)}) \) is the dimension of the mode-\( i \) subspace. The product manifold is:
\[
\mathcal{M} = S_{I_1}(r_1) \times S_{I_2}(r_2) \times S_{I_3}(r_3),
\]
where \( S_{I_i}(r_i) \) is the manifold of \( I_i \times I_i \) SSPD matrices of rank \( r_i \).

A stratified manifold is a union of smooth submanifolds (strata) of different dimensions, i.e., \( \mathcal{M} = \bigcup_{r_1, r_2, r_3} S_{I_1}(r_1) \times S_{I_2}(r_2) \times S_{I_3}(r_3) \), where each stratum corresponds to a unique rank triplet \( (r_1, r_2, r_3) \). If \( \mathcal{F} \) lies on a single smooth manifold, the rank triplet remains constant under perturbations. If the ranks vary, \( \mathcal{F} \) traverses different strata, indicating stratification.

We perturb the residual stream \( \mathbf{x} \in \mathbb{R}^{I_1 \times I_2 \times d_{{model}}} \) with Gaussian noise \( \mathcal{N}(0, \sigma^2) \), producing \( \mathbf{x}' = \mathbf{x} + \eta \), where \( \eta \sim \mathcal{N}(0, \sigma^2) \) and \( \sigma \in \{0, 0.02, \dots, 10\} \). The SAE encodes \( \mathbf{x}' \) into \( \mathcal{F}' \), with unfolding \( F'^{(i)} \). The SSPD matrix becomes:
\[
S'^{(i)} = F'^{(i)} F'^{(i)^T} + \epsilon I.
\]

Assume \( F^{(i)} \) has rank \( r_i \). Noise \( \eta \) perturbs the column space of \( F^{(i)} \), potentially increasing \( \text{rank}(F'^{(i)}) \) by activating additional linearly independent directions. For the feature mode (\( i=3 \)), \( F^{(3)} \in \mathbb{R}^{I_3 \times (I_1 I_2)} \) represents feature vectors. Noise may activate sparse features in the SAE, increasing the number of active directions, thus:
\[
r_3' = \text{rank}(F'^{(3)}) \geq r_3.
\]

In other word, the variability of \( r_3 \) under noise reflects transitions across submanifolds \( S_{I_3}(r_3) \), confirming that the latent space is a stratified manifold.

\FloatBarrier
\section{Experimental Details and Results}\label{app:Experimental Details and Results}

\subsection{Datasets and Configuration}\label{dataset and config}
To verify the stratification of the manifolds representing concepts in the latent space of SAEs, we tested the SAEMA on SAEs pre-trained on the residual streams of three different language models. The concepts tested were "Months", "Weekdays", "Chemical Elements", "Alphabet", "Color", "Constellations", "Number", "Phonetic Symbol", and "Planets", and the noise levels ranging from 0.0, 0.02, 0.05, 0.1, 0.2, 0.5, 1.0, 2.0, 5.0, to 10.0. To mitigate dead activations and ensure as many features as possible responded, we designed diverse prompts for each concept, Table~\ref{tab:prompts_wrapped} shows the prompts of "Months", "Weekdays", "Chemical Elements", Table~\ref{tab: config of language models} shows the basic configurations of the language models and Table~\ref{tab: config of SAEs} shows the corresponding configurations of the pretrained SAEs that loaded from SAE-Lens.

\begin{table}[h]
\centering
\caption{Configuration of the Pretrained SAEs}
\label{tab: config of SAEs}
\begin{adjustbox}{width=\textwidth}
\begin{tabular}{lccc}
\toprule
\textbf{Property} & \textbf{GPT2-Small Layer 11} & \textbf{Pythia-70M Layer 5} & \textbf{Gemma-2-2B Layer 19} \\
\midrule
Model Name & gpt2-small & EleutherAI/pythia-70m-deduped & gemma-2-2b \\
Release & gpt2-small-resid-post-v5-32k & pythia-70m-deduped-res-sm & sae\_bench\_gemma-2-2b\_topk\_width-2pow16\_date-1109 \\
SAE ID & blocks.11.hook\_resid\_post & blocks.5.hook\_resid\_post & blocks.19.hook\_resid\_post\_\_trainer\_0 \\
Hook Point & blocks.11.hook\_resid\_post & blocks.5.hook\_resid\_post & blocks.19.hook\_resid\_post \\
Context Size & 128 & 128 & 128 \\
$d_{\text{sae}}$ & 32,768 & 32,768 & 65,536 \\
$d_{\text{sae}} / d_{\text{model}}$ & 42.7 & 64.0 & 28.4 \\
Batch Size & 4096 & 4096 & 4096 \\
\bottomrule
\end{tabular}
\end{adjustbox}
\end{table}

\begin{table}[h]
\centering
\caption{Configuration of the language models corresponding to the SAEs}
\label{tab: config of language models}
\begin{adjustbox}{width=\textwidth}
\begin{tabular}{lccc}
\toprule
\textbf{Property} & \textbf{GPT2-Small Layer 11} & \textbf{Pythia-70M Layer 5} & \textbf{Gemma-2-2B Layer 19} \\
\midrule
Parameter Count & 124M & 70M & 2.2B \\
Number of Layers & 12 & 6 & 26 \\
Hidden Dimension (d\_model) & 768 & 512 & 2304 \\
Number of Attention Heads & 12 & 8 & 16 \\
Maximum Sequence Length & 1024 & 2048 & 8192 \\
Vocabulary Size & 50257 & 50304 & 256000 \\
Position Embedding Type & Learned & Rotary & Rotary \\
Pretraining Framework & Transformers (Hugging Face) & GPT-NeoX (EleutherAI) & Transformers (Hugging Face) \\
\bottomrule
\end{tabular}
\end{adjustbox}
\end{table}

\begin{table}[h]
    \centering
    \caption{Prompts of Months, Weekdays and Chemical Elements}
    \label{tab:prompts_wrapped}
    \scriptsize
    \begin{adjustbox}{width=\textwidth}
        \begin{tabular}{lp{0.8\textwidth}}
            \hline
            \textbf{Concept} & \multicolumn{1}{c}{\textbf{Prompt}} \\
            \hline
            \multirow{3}{*}{Months} &
            \texttt{"January", "February", "March", "April", "May", "June", "July", "August", "September", "October", "November", "December", "January starts the year.", "February follows January.", "March comes after February.", "April is the fourth month.", "May succeeds April.", "June marks midyear.", "July is the seventh month.", "August follows July.", "September begins the fall.", "October is the tenth month.", "November nears year-end.", "December ends the year.", "January in winter.", "February in winter.", "March in spring.", "April in spring.", "May in spring.", "June in summer.", "July in summer.", "August in summer.", "September in fall.", "October in fall.", "November in fall.", "December in winter.", "Month one is January.", "Month two is February.", "Month three is March.", "Month four is April.", "Month five is May.", "Month six is June.", "Month seven is July.", "Month eight is August.", "Month nine is September.", "Month ten is October.", "Month eleven is November.", "Month twelve is December.", "The month of January.", "The month of February.", "The month of March.", "The month of April.", "The month of May.", "The month of June.", "The month of July.", "The month of August.", "The month of September.", "The month of October.", "The month of November.", "The month of December."}
            \\ \hline
            \multirow{3}{*}{Weekdays} &
            \texttt{"Monday", "Tuesday", "Wednesday", "Thursday", "Friday", "Saturday", "Sunday", "Monday starts the week.", "Tuesday follows Monday.", "Wednesday comes after Tuesday.", "Thursday is the fourth day.", "Friday succeeds Thursday.", "Saturday marks the weekend.", "Sunday ends the week.", "Monday in workweek.", "Tuesday in workweek.", "Wednesday in workweek.", "Thursday in workweek.", "Friday in workweek.", "Saturday in weekend.", "Sunday in weekend.", "Day one is Monday.", "Day two is Tuesday.", "Day three is Wednesday.", "Day four is Thursday.", "Day five is Friday.", "Day six is Saturday.", "Day seven is Sunday.", "The day of Monday.", "The day of Tuesday.", "The day of Wednesday.", "The day of Thursday.", "The day of Friday.", "The day of Saturday.", "The day of Sunday."}
            \\ \hline
            \multirow{3}{*}{Chemical Elements} &
            \texttt{"Hydrogen", "Helium", "Lithium", "Beryllium", "Boron", "Carbon", "Nitrogen", "Oxygen", "Fluorine", "Neon", "Sodium", "Magnesium", "Hydrogen starts the periodic table.", "Helium follows Hydrogen.", "Lithium comes after Helium.", "Beryllium is the fourth element.", "Boron succeeds Beryllium.", "Carbon follows Boron.", "Nitrogen comes after Carbon.", "Oxygen succeeds Nitrogen.", "Fluorine follows Oxygen.", "Neon comes after Fluorine.", "Sodium starts the third period.", "Magnesium follows Sodium.", "Hydrogen is a non-metal.", "Helium is a noble gas.", "Lithium is an alkali metal.", "Beryllium is an alkaline earth metal.", "Boron is a metalloid.", "Carbon forms life.", "Nitrogen is in air.", "Oxygen supports life.", "Fluorine is highly reactive.", "Neon glows in signs.", "Sodium is in salt.", "Magnesium burns brightly.", "Hydrogen in period one.", "Helium in period one.", "Lithium in period two.", "Beryllium in period two.", "Boron in period two.", "Carbon in period two.", "Nitrogen in period two.", "Oxygen in period two.", "Fluorine in period two.", "Neon in period two.", "Sodium in period three.", "Magnesium in period three.", "Element one is Hydrogen.", "Element two is Helium.", "Element three is Lithium.", "Element four is Beryllium.", "Element five is Boron.", "Element six is Carbon.", "Element seven is Nitrogen.", "Element eight is Oxygen.", "Element nine is Fluorine.", "Element ten is Neon.", "Element eleven is Sodium.", "Element twelve is Magnesium.", "The element of Hydrogen.", "The element of Helium.", "The element of Lithium.", "The element of Beryllium.", "The element of Boron.", "The element of Carbon.", "The element of Nitrogen.", "The element of Oxygen.", "The element of Fluorine.", "The element of Neon.", "The element of Sodium.", "The element of Magnesium."}
            \\ \hline
        \end{tabular}
    \end{adjustbox}
\end{table}

\subsection{Experiment Results of Case 1}\label{exp:case1}
 The variation of the rank of the SSPD matrices and Average Geodesic Distance (AGD) corresponding to the modal matrices unfolded along the three dimensions with different noise levels is shown in Table \ref{tab: rank & AGD vs noise(Months+days+elements)} and \ref{tab: rank & AGD vs noise (last 6 concepts)}.

\begin{table}[h]
\centering
\caption{Ranks of the SSPD matrix ($r_1$, $r_2$, $r_3$) and AGD values for pre-trained SAEs across three concepts (\textbf{Months}, \textbf{Weekdays}, \textbf{Chemical Elements}) under varying noise levels.}
\label{tab: rank & AGD vs noise(Months+days+elements)}
\begin{adjustbox}{width=\textwidth}
\begin{tabular}{lccccccccccccc}
\toprule
\multirow{2}{*}{Concept} & \multirow{2}{*}{Noise Level} & \multicolumn{4}{c}{$SAE_{GPT2-Small\ Layer\ 11}$} & \multicolumn{4}{c}{$SAE_{Pythia-70M\ Layer\ 5}$} & \multicolumn{4}{c}{$SAE_{Gemma-2-2B\ Layer\ 19}$} \\ \cmidrule(lr){3-6} \cmidrule(lr){7-10} \cmidrule(lr){11-14}
& & $r_1$ & $r_2$ & $r_3$ & AGD & $r_1$ & $r_2$ & $r_3$ & AGD & $r_1$ & $r_2$ & $r_3$ & AGD \\ \midrule
\multirow{10}{*}{Months} & 0.00 & 45 & 6 & 146 & 11.97 & 45 & 6 & 216 & 8.33 & 45 & 5 & 118 & 1319.10 \\
& 0.02 & 45 & 6 & 147 & 11.97 & 45 & 6 & 251 & 8.34 & 45 & 5 & 119 & 1319.24 \\
& 0.05 & 45 & 6 & 149 & 11.97 & 45 & 6 & 290 & 8.38 & 45 & 5 & 120 & 1319.53 \\
& 0.10 & 45 & 6 & 155 & 11.98 & 45 & 6 & 309 & 8.46 & 45 & 5 & 125 & 1319.79 \\
& 0.20 & 45 & 6 & 165 & 11.98 & 45 & 6 & 346 & 8.69 & 45 & 5 & 129 & 1320.09 \\
& 0.50 & 45 & 6 & 207 & 11.98 & 45 & 6 & 360 & 12.08 & 45 & 5 & 144 & 1320.87 \\
& 1.00 & 45 & 6 & 251 & 12.00 & 45 & 6 & 360 & 36.58 & 45 & 5 & 162 & 1321.62 \\
& 2.00 & 45 & 6 & 305 & 12.02 & 45 & 6 & 359 & 157.30 & 45 & 5 & 174 & 1322.93 \\
& 5.00 & 45 & 6 & 360 & 12.21 & 45 & 6 & 360 & 572.87 & 45 & 5 & 196 & 1327.87 \\
& 10.00 & 45 & 6 & 360 & 13.46 & 45 & 6 & 360 & 1262.54 & 45 & 5 & 237 & 1355.26 \\
\midrule
\multirow{10}{*}{Weekdays} & 0.00 & 26 & 5 & 88 & 11.87 & 26 & 5 & 124 & 9.41 & 26 & 5 & 82 & 1304.22 \\
& 0.02 & 26 & 5 & 89 & 11.87 & 26 & 5 & 159 & 9.43 & 26 & 5 & 82 & 1304.28 \\
& 0.05 & 26 & 5 & 93 & 11.88 & 26 & 5 & 171 & 9.43 & 26 & 5 & 85 & 1304.61 \\
& 0.10 & 26 & 5 & 100 & 11.88 & 26 & 5 & 181 & 9.55 & 26 & 5 & 89 & 1304.88 \\
& 0.20 & 26 & 5 & 108 & 11.88 & 26 & 5 & 183 & 9.77 & 26 & 5 & 95 & 1305.37 \\
& 0.50 & 26 & 5 & 141 & 11.89 & 26 & 5 & 183 & 12.95 & 26 & 5 & 107 & 1305.97 \\
& 1.00 & 26 & 5 & 163 & 11.89 & 26 & 5 & 183 & 42.71 & 26 & 5 & 114 & 1306.61 \\
& 2.00 & 26 & 5 & 182 & 11.95 & 26 & 5 & 183 & 149.61 & 26 & 5 & 120 & 1307.61 \\
& 5.00 & 26 & 5 & 183 & 12.32 & 26 & 5 & 183 & 570.94 & 26 & 5 & 138 & 1317.50 \\
& 10.00 & 26 & 5 & 183 & 13.56 & 26 & 5 & 183 & 1348.77 & 26 & 5 & 165 & 1338.49 \\
\midrule
\multirow{10}{*}{Chemical Elements} & 0.00 & 54 & 9 & 241 & 10.70 & 54 & 8 & 375 & 9.46 & 54 & 6 & 179 & 1171.65 \\
& 0.02 & 54 & 9 & 245 & 10.70 & 54 & 8 & 397 & 9.47 & 54 & 6 & 180 & 1171.74 \\
& 0.05 & 54 & 9 & 250 & 10.70 & 54 & 8 & 469 & 9.49 & 54 & 6 & 180 & 1171.90 \\
& 0.10 & 54 & 9 & 266 & 10.70 & 54 & 8 & 521 & 9.54 & 54 & 6 & 185 & 1172.07 \\
& 0.20 & 54 & 9 & 285 & 10.70 & 54 & 8 & 555 & 9.81 & 54 & 6 & 189 & 1172.29 \\
& 0.50 & 54 & 9 & 330 & 10.70 & 54 & 8 & 591 & 13.44 & 54 & 6 & 201 & 1172.82 \\
& 1.00 & 54 & 9 & 378 & 10.72 & 54 & 8 & 594 & 40.58 & 54 & 6 & 222 & 1173.63 \\
& 2.00 & 54 & 9 & 419 & 10.77 & 54 & 8 & 592 & 153.71 & 54 & 6 & 240 & 1174.72 \\
& 5.00 & 54 & 9 & 551 & 11.17 & 54 & 8 & 593 & 578.41 & 54 & 6 & 261 & 1183.68 \\
& 10.00 & 54 & 9 & 645 & 12.86 & 54 & 8 & 592 & 1309.70 & 54 & 6 & 320 & 1203.08 \\
\bottomrule
\end{tabular}
\end{adjustbox}
\end{table}

\begin{table}[h]
\centering
\caption{Ranks of the SSPD matrix ($r_1$, $r_2$, $r_3$) and AGD values for pre-trained SAEs across six concepts (\textbf{Alphabet}, \textbf{Color}, \textbf{Constellations}, \textbf{Number}, \textbf{Phonetic Symbol}, \textbf{Planets}) under varying noise levels.}
\label{tab: rank & AGD vs noise (last 6 concepts)}
\begin{adjustbox}{width=\textwidth}
\begin{tabular}{lccccccccccccc}
\toprule
\multirow{2}{*}{Concept} & \multirow{2}{*}{Noise Level} & \multicolumn{4}{c}{$SAE_{GPT2-Small\ Layer\ 11}$} & \multicolumn{4}{c}{$SAE_{Pythia-70M\ Layer\ 5}$} & \multicolumn{4}{c}{$SAE_{Gemma-2-2B\ Layer\ 19}$} \\ \cmidrule(lr){3-6} \cmidrule(lr){7-10} \cmidrule(lr){11-14}
& & $r_1$ & $r_2$ & $r_3$ & AGD & $r_1$ & $r_2$ & $r_3$ & AGD & $r_1$ & $r_2$ & $r_3$ & AGD \\ \midrule
\multirow{10}{*}{Alphabet} & 0.00 & 97 & 6 & 312 & 11.17 & 97 & 6 & 468 & 8.63 & 97 & 6 & 294 & 631.31 \\
& 0.02 & 97 & 6 & 314 & 11.21 & 97 & 6 & 498 & 8.67 & 97 & 6 & 294 & 631.31 \\
& 0.05 & 97 & 6 & 316 & 11.15 & 97 & 6 & 572 & 8.66 & 97 & 6 & 294 & 631.32 \\
& 0.10 & 97 & 6 & 321 & 11.09 & 97 & 6 & 633 & 8.69 & 97 & 6 & 294 & 631.32 \\
& 0.20 & 97 & 6 & 331 & 11.24 & 97 & 6 & 705 & 9.01 & 97 & 6 & 297 & 632.21 \\
& 0.50 & 97 & 6 & 383 & 11.28 & 97 & 6 & 778 & 12.45 & 97 & 6 & 298 & 632.21 \\
& 1.00 & 97 & 6 & 453 & 11.31 & 97 & 6 & 780 & 38.04 & 97 & 6 & 368 & 634.57 \\
& 2.00 & 97 & 6 & 563 & 11.41 & 97 & 6 & 773 & 163.13 & 97 & 6 & 387 & 644.17 \\
& 5.00 & 97 & 6 & 732 & 11.47 & 97 & 6 & 772 & 608.19 & 97 & 6 & 490 & 650.92 \\
& 10.00 & 97 & 6 & 864 & 12.82 & 97 & 6 & 776 & 1334.49 & 97 & 6 & 633 & 656.45 \\
\midrule
\multirow{10}{*}{Color} & 0.00 & 26 & 5 & 91 & 11.99 & 26 & 5 & 116 & 9.05 & 26 & 5 & 96 & 630.61 \\
& 0.02 & 26 & 5 & 93 & 11.99 & 26 & 5 & 149 & 9.06 & 26 & 5 & 95 & 630.61 \\
& 0.05 & 26 & 5 & 97 & 11.99 & 26 & 5 & 168 & 9.07 & 26 & 5 & 95 & 630.62 \\
& 0.10 & 26 & 5 & 105 & 11.99 & 26 & 5 & 177 & 9.11 & 26 & 5 & 96 & 630.61 \\
& 0.20 & 26 & 5 & 115 & 11.99 & 26 & 5 & 181 & 9.45 & 26 & 5 & 96 & 630.68 \\
& 0.50 & 26 & 5 & 142 & 12.00 & 26 & 5 & 183 & 12.76 & 26 & 5 & 101 & 630.81 \\
& 1.00 & 26 & 5 & 165 & 12.01 & 26 & 5 & 184 & 39.83 & 26 & 5 & 111 & 631.33 \\
& 2.00 & 26 & 5 & 181 & 12.07 & 26 & 5 & 183 & 149.47 & 26 & 5 & 132 & 632.19 \\
& 5.00 & 26 & 5 & 183 & 12.44 & 26 & 5 & 183 & 609.77 & 26 & 5 & 173 & 637.02 \\
& 10.00 & 26 & 5 & 183 & 13.87 & 26 & 5 & 183 & 1360.58 & 26 & 5 & 182 & 658.20 \\
\midrule
\multirow{10}{*}{Constellations} & 0.00 & 54 & 7 & 231 & 11.38 & 54 & 7 & 345 & 9.32 & 54 & 5 & 164 & 622.75 \\
& 0.02 & 54 & 7 & 237 & 11.38 & 54 & 7 & 368 & 9.32 & 54 & 5 & 165 & 622.77 \\
& 0.05 & 54 & 7 & 244 & 11.38 & 54 & 7 & 443 & 9.34 & 54 & 5 & 163 & 622.78 \\
& 0.10 & 54 & 7 & 252 & 11.38 & 54 & 7 & 478 & 9.41 & 54 & 5 & 164 & 622.78 \\
& 0.20 & 54 & 7 & 269 & 11.38 & 54 & 7 & 509 & 9.69 & 54 & 5 & 165 & 622.77 \\
& 0.50 & 54 & 7 & 315 & 11.39 & 54 & 7 & 539 & 12.92 & 54 & 5 & 168 & 622.85 \\
& 1.00 & 54 & 7 & 363 & 11.39 & 54 & 7 & 540 & 39.70 & 54 & 5 & 180 & 623.08 \\
& 2.00 & 54 & 7 & 406 & 11.44 & 54 & 7 & 540 & 154.04 & 54 & 5 & 207 & 624.13 \\
& 5.00 & 54 & 7 & 501 & 11.81 & 54 & 7 & 539 & 617.11 & 54 & 5 & 322 & 629.99 \\
& 10.00 & 54 & 7 & 540 & 13.43 & 54 & 7 & 540 & 1322.80 & 54 & 5 & 361 & 649.63 \\
\midrule
\multirow{10}{*}{Number} & 0.00 & 41 & 5 & 132 & 11.72 & 41 & 5 & 184 & 8.87 & 41 & 5 & 132 & 625.66 \\
& 0.02 & 41 & 5 & 132 & 11.72 & 41 & 5 & 215 & 8.88 & 41 & 5 & 132 & 625.67 \\
& 0.05 & 41 & 5 & 135 & 11.72 & 41 & 5 & 251 & 8.91 & 41 & 5 & 132 & 625.66 \\
& 0.10 & 41 & 5 & 144 & 11.72 & 41 & 5 & 265 & 8.96 & 41 & 5 & 131 & 625.70 \\
& 0.20 & 41 & 5 & 152 & 11.73 & 41 & 5 & 283 & 9.22 & 41 & 5 & 135 & 625.71 \\
& 0.50 & 41 & 5 & 187 & 11.74 & 41 & 5 & 288 & 12.39 & 41 & 5 & 138 & 625.85 \\
& 1.00 & 41 & 5 & 229 & 11.75 & 41 & 5 & 289 & 37.19 & 41 & 5 & 149 & 626.41 \\
& 2.00 & 41 & 5 & 264 & 11.81 & 41 & 5 & 288 & 145.61 & 41 & 5 & 174 & 627.09 \\
& 5.00 & 41 & 5 & 288 & 12.20 & 41 & 5 & 288 & 588.49 & 41 & 5 & 261 & 632.37 \\
& 10.00 & 41 & 5 & 288 & 13.58 & 41 & 5 & 288 & 1317.34 & 41 & 5 & 282 & 652.09 \\
\midrule
\multirow{10}{*}{Phonetic Symbol} & 0.00 & 42 & 8 & 177 & 10.55 & 42 & 9 & 246 & 14.56 & 42 & 7 & 168 & 487.41 \\
& 0.02 & 42 & 8 & 179 & 10.55 & 42 & 9 & 250 & 14.57 & 42 & 7 & 169 & 487.42 \\
& 0.05 & 42 & 8 & 180 & 10.55 & 42 & 9 & 281 & 14.58 & 42 & 7 & 169 & 487.43 \\
& 0.10 & 42 & 8 & 190 & 10.55 & 42 & 9 & 349 & 14.65 & 42 & 7 & 170 & 487.45 \\
& 0.20 & 42 & 8 & 213 & 10.56 & 42 & 9 & 431 & 14.93 & 42 & 7 & 170 & 487.46 \\
& 0.50 & 42 & 8 & 256 & 10.56 & 42 & 9 & 500 & 18.19 & 42 & 7 & 180 & 487.68 \\
& 1.00 & 42 & 8 & 299 & 10.58 & 42 & 9 & 504 & 44.25 & 42 & 7 & 192 & 488.24 \\
& 2.00 & 42 & 8 & 348 & 10.63 & 42 & 9 & 502 & 149.26 & 42 & 7 & 224 & 489.80 \\
& 5.00 & 42 & 8 & 435 & 10.95 & 42 & 9 & 504 & 608.31 & 42 & 7 & 374 & 498.49 \\
& 10.00 & 42 & 8 & 462 & 12.53 & 42 & 9 & 503 & 1369.46 & 42 & 7 & 408 & 525.42 \\
\midrule
\multirow{10}{*}{Planets} & 0.00 & 30 & 8 & 121 & 10.66 & 30 & 8 & 220 & 8.59 & 30 & 6 & 113 & 563.82 \\
& 0.02 & 30 & 8 & 128 & 10.66 & 30 & 8 & 255 & 8.59 & 30 & 6 & 113 & 563.83 \\
& 0.05 & 30 & 8 & 132 & 10.66 & 30 & 8 & 305 & 8.62 & 30 & 6 & 114 & 563.82 \\
& 0.10 & 30 & 8 & 145 & 10.66 & 30 & 8 & 319 & 8.69 & 30 & 6 & 113 & 563.81 \\
& 0.20 & 30 & 8 & 154 & 10.66 & 30 & 8 & 327 & 9.03 & 30 & 6 & 114 & 563.84 \\
& 0.50 & 30 & 8 & 189 & 10.67 & 30 & 8 & 330 & 12.95 & 30 & 6 & 119 & 564.08 \\
& 1.00 & 30 & 8 & 219 & 10.69 & 30 & 8 & 330 & 39.84 & 30 & 6 & 129 & 564.13 \\
& 2.00 & 30 & 8 & 261 & 10.73 & 30 & 8 & 329 & 164.59 & 30 & 6 & 156 & 566.21 \\
& 5.00 & 30 & 8 & 325 & 11.15 & 30 & 8 & 330 & 613.29 & 30 & 6 & 228 & 572.72 \\
& 10.00 & 30 & 8 & 330 & 12.66 & 30 & 8 & 329 & 1290.54 & 30 & 6 & 237 & 595.88 \\
\bottomrule
\end{tabular}
\end{adjustbox}
\end{table}

\FloatBarrier

\subsection{Discussion of the results of the experiment in Case 2}\label{app: discussion_case2}

Liu et al.~\cite{52} pointed out that the superposition phenomenon is a key mechanism behind scaling laws—that is, the larger the model scale, the more pronounced the superposition phenomenon. As the smallest of the three language models, Pythia 70M exhibits the weakest superposition, indicating that the overlap in local features is not severe. Moreover, as shown in Table~\ref{tab: config of SAEs}, the expansion factor of $SAE\ Pythia\ 70M\ Layer\ 5$ is 64.0, indicating the highest level of sparsity. Therefore, among the three concepts represented by $SAE\ Pythia\ 70M\ Layer\ 5$, the number of connected components decreases for two of them, which may be due to the weaker feature overlap within its local representations, resulting in a separation trend that is less dominant than the merging of similar semantics.  

If the degree of feature overlap in local representations is taken as an indirect indicator of decoupling, then both the increase in $Avg.\ ID$ accompanied by an increase in $Betti\ 0$ in large-scale models and the increase in $Avg.\ ID$ accompanied by a decrease in $Betti\ 0$ in small-scale models could indirectly suggest a reduction in feature overlap—that is, the successful separation of monosemantic features.  

\FloatBarrier

\subsection{Experiment Results of Case 3}\label{app:exp_resu_case3}
\subsubsection{Visualization Results of the Optimization Process based on Equation~\ref{eq:gw_loss}}\label{exp:case3}
\begin{figure}[h]
    \centering
    \includegraphics[width=1.0\linewidth]{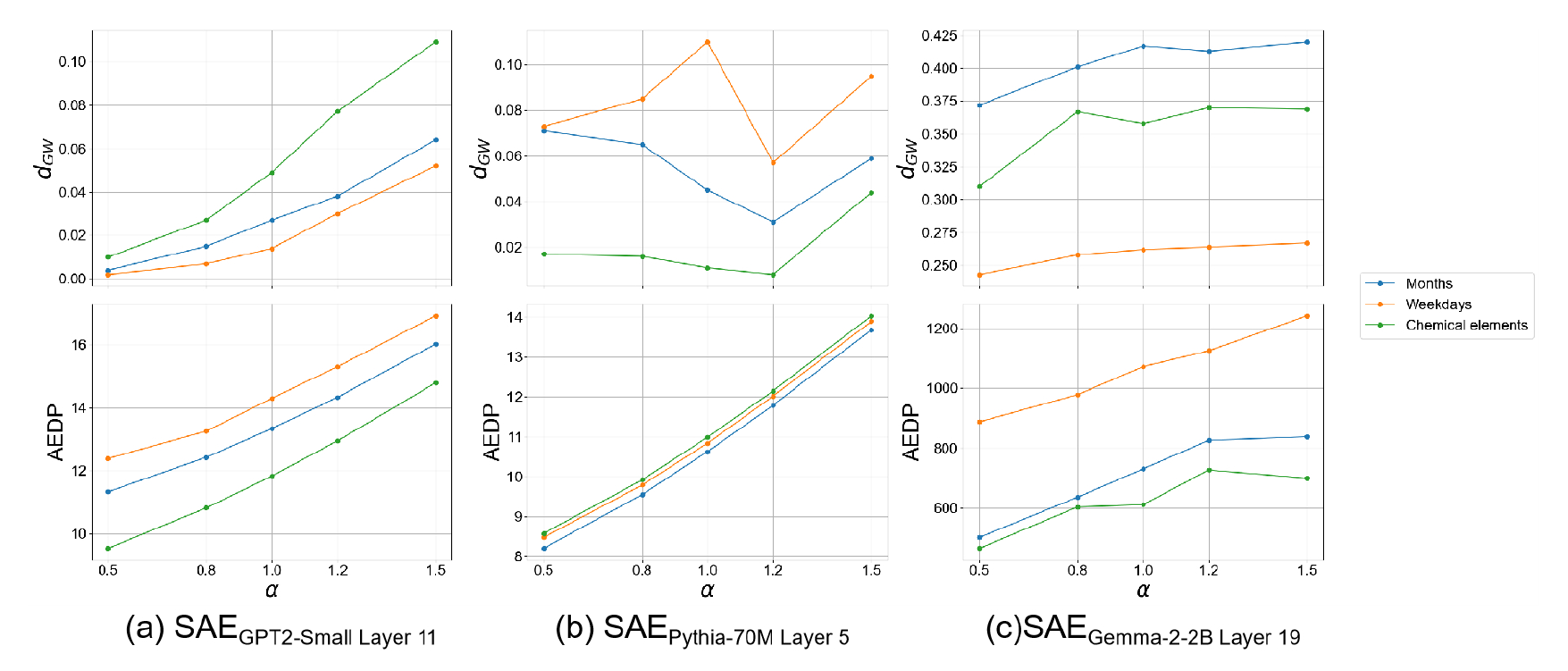}
    \caption{Variability of $d_{GW}$ and AEDP with increasing $\alpha$ during the optimization of Equation~\ref{eq:gw_loss}.}
    \label{fig:case3_dGW_AEDP_vs_Alpha}
\end{figure}

\begin{figure}[h]
    \centering
    \includegraphics[width=1.0\linewidth]{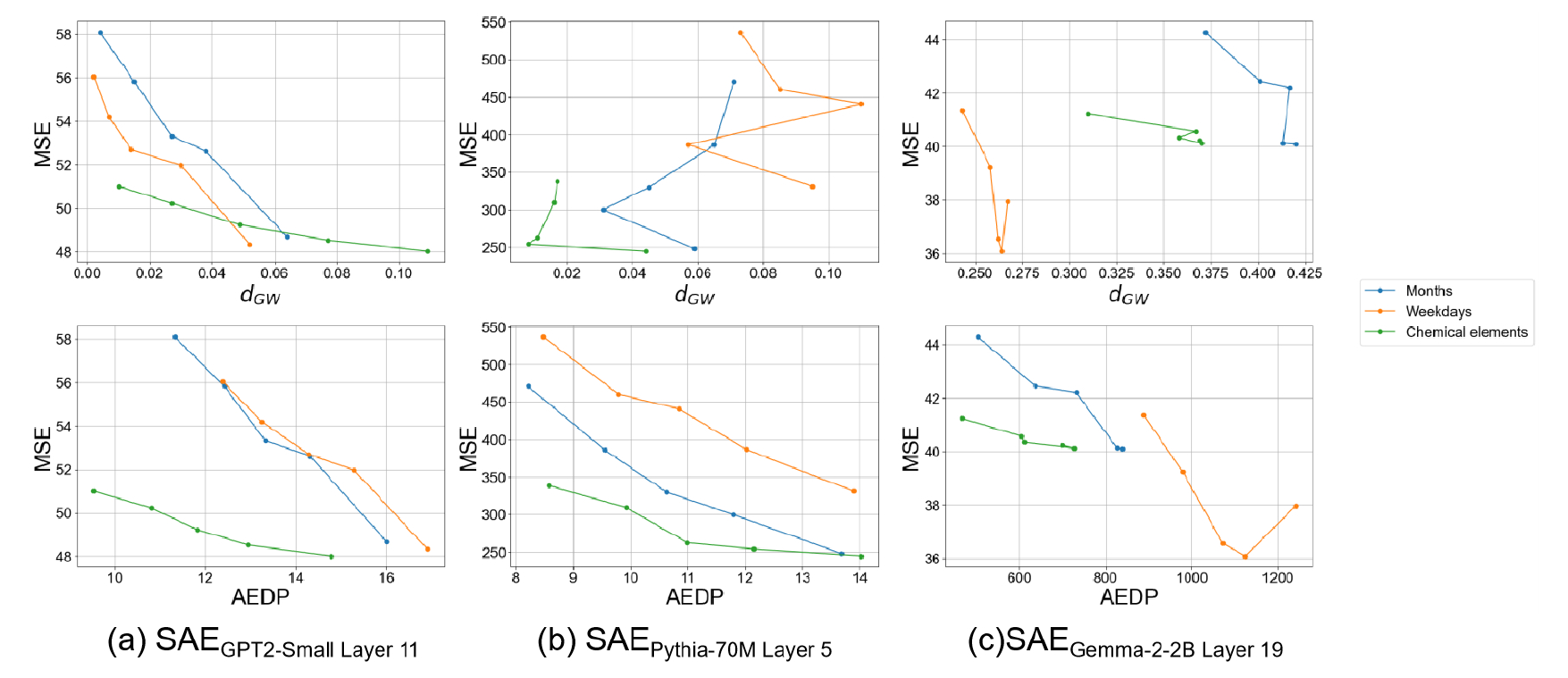}
    \caption{Variability of $d_{GW}$ and AEDP with MSE for different $\alpha$ during the optimization of Equation~\ref{eq:gw_loss}.}
    \label{fig:case3_MSE_vs_dGW_AEDP}
\end{figure}

\subsubsection{Experiment Details and Results of the Optimization Process based on Equation~\ref{eq: AEDP_loss}}\label{case3.2}
To validate the causal relationship between the separability of the proposed local representations and the reconstruction performance of SAE, we modified the loss function in Equation~\ref{eq:gw_loss} by replacing \(d_{GW}\) with \(AEDP^{-1}\), as shown in Equation~\ref{eq: AEDP_loss}. The implementation of the optimization process remains the same as in Equation~\ref{eq:gw_loss}:

\begin{equation}
L = AEDP^{-1} + \lambda_{MSE} MSE(x,\hat{x})
\label{eq: AEDP_loss}
\end{equation}

\begin{figure}[h]
    \centering
    \includegraphics[width=1.0\linewidth]{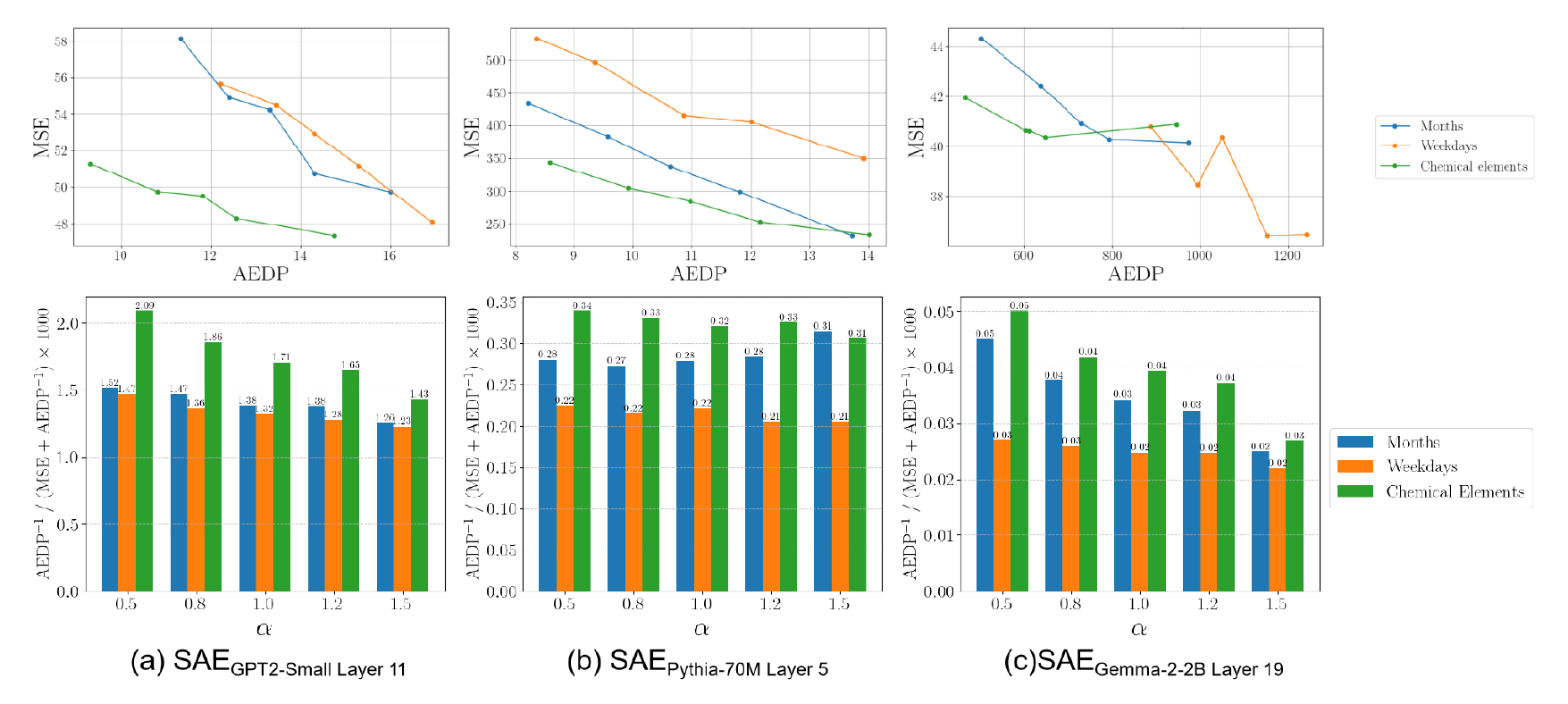}
    \caption{First column: Variability of MSE with AEDP as $\alpha$ increases during the optimization of Equation~\ref{eq: AEDP_loss}; Second column: Contribution of the $AEDP^{-1}$ term to the Total Loss during the optimization of Equation~\ref{eq: AEDP_loss}.}
    \label{fig:Case3.2_AEDPvsMSE_contribution}
\end{figure}

Table~\ref{tab:case3.2} presents the changes in MSE and AEDP for different \(\alpha\) values. The first column of Figure~\ref{fig:Case3.2_AEDPvsMSE_contribution} illustrates how MSE varies with AEDP, which aligns with the trend observed in the second column of Figure~\ref{fig:case3_MSE_vs_dGW_AEDP} (i.e., Equation~\ref{eq:gw_loss})—MSE decreases as AEDP increases. Additionally, the second column of Figure~\ref{fig:Case3.2_AEDPvsMSE_contribution} shows the contribution of \(AEDP^{-1}\) to Equation~\ref{eq: AEDP_loss} during the optimization process. Even when magnified 1000 times, its proportion remains negligible. Based on these observations, we conclude that MSE consistently dominates the optimization process, and its reduction is directly caused by the increase in AEDP. Combined with the experimental results of the optimization based on Equation~\ref{eq:gw_loss}, we conclude that the enhanced separability among local representations directly leads to improved reconstruction performance in SAE.

\begin{table}[h]
\centering
\caption{Variations of MSE and AEDP corresponding to different translation step sizes ($\alpha$) during optimization based on Equation~\ref{eq: AEDP_loss}.}
\label{tab:case3.2}
\begin{adjustbox}{width=\textwidth}
\begin{tabular}{lccccccccccccc}
\toprule
\multirow{2}{*}{Concept} & \multirow{2}{*}{$\alpha$} & \multicolumn{4}{c}{$SAE_{GPT2-Small\ Layer\ 11}$} & \multicolumn{4}{c}{$SAE_{Pythia-70M\ Layer\ 5}$} & \multicolumn{4}{c}{$SAE_{Gemma-2-2B\ Layer\ 19}$} \\ \cmidrule(lr){3-6} \cmidrule(lr){7-10} \cmidrule(lr){11-14}
& & MSE$\downarrow$ & Orig. AEDP & AEDP$\uparrow$ & $AEDP^{-1}\downarrow$ & MSE$\downarrow$ & Orig. AEDP & AEDP$\uparrow$ & $AEDP^{-1}\downarrow$ & MSE$\downarrow$ & Orig. AEDP & AEDP$\uparrow$ & $AEDP^{-1}\downarrow$ \\ \midrule
\multirow{5}{*}{Months} & 0.5 & 58.16 & 10.50 & 11.32 & 0.0883 & 434.09 & 7.19 & 8.21 & 0.1218 & 44.33 & 358.97 & 503.24 & 0.0020 \\
& 0.8 & 54.90 & 10.50 & 12.40 & 0.0806 & 383.16 & 7.19 & 9.56 & 0.1046 & 42.42 & 358.97 & 637.58 & 0.0016 \\
& 1.0 & 54.24 & 10.50 & 13.32 & 0.0751 & 337.16 & 7.19 & 10.63 & 0.0941 & 40.91 & 358.97 & 730.67 & 0.0014 \\
& 1.2 & 50.74 & 10.50 & 14.31 & 0.0699 & 298.28 & 7.19 & 11.81 & 0.0847 & 40.28 & 358.97 & 793.36 & 0.0013 \\
& 1.5 & 49.72 & 10.50 & 16.00 & 0.0625 & 231.63 & 7.19 & 13.72 & 0.0729 & 40.14 & 358.97 & 973.32 & 0.0010 \\
\midrule
\multirow{5}{*}{Weekdays} & 0.5 & 55.67 & 11.58 & 12.21 & 0.0819 & 533.29 & 7.46 & 8.35 & 0.1198 & 40.78 & 774.53 & 888.47 & 0.0011 \\
& 0.8 & 54.48 & 11.58 & 13.46 & 0.0743 & 496.42 & 7.46 & 9.35 & 0.1070 & 38.45 & 774.53 & 995.49 & 0.0010 \\
& 1.0 & 52.92 & 11.58 & 14.31 & 0.0699 & 415.42 & 7.46 & 10.86 & 0.0921 & 40.36 & 774.53 & 1050.32 & 0.0010 \\
& 1.2 & 51.17 & 11.58 & 15.28 & 0.0654 & 405.19 & 7.46 & 12.01 & 0.0833 & 36.41 & 774.53 & 1152.77 & 0.0009 \\
& 1.5 & 48.07 & 11.58 & 16.91 & 0.0591 & 349.96 & 7.46 & 13.91 & 0.0719 & 36.45 & 774.53 & 1241.62 & 0.0008 \\
\midrule
\multirow{5}{*}{Chemical Elements} & 0.5 & 51.27 & 8.49 & 9.32 & 0.1073 & 343.42 & 7.53 & 8.58 & 0.1166 & 41.94 & 318.12 & 467.35 & 0.0021 \\
& 0.8 & 49.74 & 8.49 & 10.81 & 0.0925 & 304.37 & 7.53 & 9.92 & 0.1008 & 40.63 & 318.12 & 604.37 & 0.0017 \\
& 1.0 & 49.50 & 8.49 & 11.82 & 0.0846 & 284.00 & 7.53 & 10.97 & 0.0912 & 40.62 & 318.12 & 612.15 & 0.0016 \\
& 1.2 & 48.29 & 8.49 & 12.55 & 0.0797 & 252.52 & 7.53 & 12.16 & 0.0822 & 40.35 & 318.12 & 648.64 & 0.0015 \\
& 1.5 & 47.33 & 8.49 & 14.75 & 0.0678 & 232.69 & 7.53 & 14.01 & 0.0713 & 40.87 & 318.12 & 945.88 & 0.0011 \\
\bottomrule
\end{tabular}
\end{adjustbox}
\end{table}

\subsubsection{Discussion of the results of the experiment in Case 3}\label{app: discussion_case3}

The reconstruction error of the three pretrained SAEs exhibits a certain degree of negative correlation with the scale of the corresponding language models. In other word, the larger the language model, the better the reconstruction performance. This stems from the conflict between sparsity and reconstruction performance. The activation frequency of $SAE\ Pythia\ 70M\ Layer\ 5$ is approximately 0.000026\%, that of $SAE\ GPT2-Small\ Layer\ 11$ is approximately 0.0039\%, and the activation proportion of $SAE\ Gemma-2-2B\ Layer\ 19$ is approximately 0.03\%, indicating that higher sparsity leads to poorer reconstruction performance. The expansion factor in Table~\ref{tab: config of SAEs} serve as indicators to trade off sparsity and representational performance. Although larger expansion factors may slightly degrade reconstruction performance, our finding regarding the causality between the separability of local representations and reconstruction performance suggests that larger expansion factors also offer greater potential for improving reconstruction performance.

\FloatBarrier

\section{Evaluation Metrics}\label{intro_eva_metx}

\paragraph{Average Intrinsic Dimensionality (Avg. ID)~\cite{50}}
Measures the effective dimensionality of local representations, reflecting degrees of freedom introduced by sparse encoding, computed via PCA eigenvalues.

\[
\text{ID}(\mathcal{M}_c^{(k)}) = \sum_{j=1}^d \mathbb{1}\left( \frac{\lambda_j}{\sum_i \lambda_i} > \tau_{\text{dim}} \right),
\quad
\text{Avg. ID} = \frac{1}{K} \sum_{k=1}^K \text{ID}(\mathcal{M}_c^{(k)}).
\]
where \(\lambda_j\) are eigenvalues of the cluster \(\mathcal{C}_k\) covariance matrix, \(\tau_{\text{dim}} \approx 0.01\), and \(K\) is the number of clusters.

\paragraph{Betti 0~\cite{49}}
Counts the connected components in a local representation, assessing the topological impact of sparse encoding.

\[
\text{Betti}_0(\mathcal{M}_c^{(k)}) = \left| \{ H_0 \text{ bars with length} > \tau_{\text{pers}} \} \right|,
\quad
\text{AvgBetti}_0 = \frac{1}{K} \sum_{k=1}^K \text{Betti}_0(\mathcal{M}_c^{(k)}).
\]
where \(H_0\) is the 0-dimensional persistence diagram, and \(\tau_{\text{pers}} \approx 0.1\).

\paragraph{Minimum Spanning Tree Weight (MSTW)~\cite{48}}
Quantifies the dispersion of cluster centroids in the global representation, computed as the sum of edges in the minimum spanning tree.

\[
\text{MSTW}(\mathcal{G}_c) = \sum_{e \in E} \|\mathbf{c}_i - \mathbf{c}_j\|_2.
\]
where \(\mathbf{c}_k\) is the centroid of cluster \(\mathcal{C}_k\), and \(E\) is the edge set of the minimum spanning tree.

\paragraph{Average Geodesic Distance (AGD)~\cite{47}}
Measures geometric differences between SSPD matrices on the manifold, validating representation stability.

\[
d_{\text{geo}}(S_m, S_n) = \sqrt{\text{tr}(S_m) + \text{tr}(S_n) - 2 \text{tr}\left( (S_m^{1/2} S_n S_m^{1/2})^{1/2} \right)},
\quad
\text{AGD} = \frac{1}{N(N-1)/2} \sum_{m < n} d_{\text{geo}}(S_m, S_n).
\]
where \(S_m\) are SSPD matrices, and \(N\) is the number of samples.

\paragraph{Procrustes Disparity}
Compares geometric rearrangement of centroid sets before and after sparse encoding, using Procrustes analysis.

\[
\text{Procrustes Disparity} = \min_{R,T} \sum_{k=1}^K \|\mathbf{c}_k^{\text{resid}} - (R \mathbf{c}_k^{\text{latent}} + T)\|_2^2.
\]
where \(\mathbf{c}_k^{\text{resid}}\) and \(\mathbf{c}_k^{\text{latent}}\) are centroids of residual stream and latent tensor, respectively, and \(R\), \(T\) are rotation and translation matrices.

\textbf{The Average Euclidean Distance between Pairs (AEDP)} of cluster centers is used to quantify the separability between different local representations.

\[
\text{AEDP} = \frac{2}{K(K-1)} \sum_{i < j} \|\mathbf{c}_i - \mathbf{c}_j\|_2
\]

where \(\mathbf{c}_i\) and \(\mathbf{c}_j\) are the centers of clusters \(i\) and \(j\), \(K\) is the number of clusters, and \(\|\cdot\|_2\) denotes the Euclidean norm.

\textbf{Mean Squared Error (MSE)~\cite{51}} is introduced to quantify the reconstruction performance of SAEs.

\[
\text{MSE} = \frac{1}{N} \sum_{i=1}^{N} \left( \hat{x}_{interv} - x \right)^2
\]

where $\hat{x}_{interv}$ is the reconstructed residual generated by decoding the intervened latent representation through the SAE, and $x$ is the original residual stream.

\FloatBarrier

\section{Limitations}\label{limitation}
Although we have confirmed that the representation structure of SAEs pre-trained on the residual stream of language models is stratified, the extraction of activations depends on prompts related to the concept. Despite designing prompts to be as diverse and information-dense as possible, there is still no guarantee that all representations of the concept have been captured (which remains an open problem). Therefore, the rigor of the conclusions needs to be further examined. Additionally, regularization serve as a critical component of SAEs, serve as a role in controlling sparsity. However, in this work, we do not included as a controlled variable. Thus, the differences in representation structure caused by variations in regularization require further exploration.
\FloatBarrier

\section{Impact Statement}\label{broader impact}
Our paper explains how Sparse Autoencoders (SAEs) organize the representations of activation vectors in language models from the perspective of representation geometry. It reveals the mechanism by which SAEs achieve feature disentanglement from the representation level for the first time. In addition, demonstrates a causal relationship between the separability of local representations and reconstruction performance, advancing research in mechanistic interpretability. As a foundational research, this work provides a reference for developing new neural network decomposition paradigms and optimizing SAEs, without direct societal impact.

\FloatBarrier

\section{Reproducibility}\label{reproducibility}

\begin{algorithm}
\caption{Pseudocode for Case 1}
\begin{algorithmic}[1]
\State \textbf{Input:}
\State \quad Models: $\mathcal{M} = \{ \text{GPT2-Small L11}, \text{Pythia-70M L5}, \text{Gemma-2-2B L19} \}$
\State \quad Concepts: $\mathcal{C} = \{ \text{Months}, \text{Weekdays}, \text{Elements}, \ldots \}$
\State \quad Noise levels: $\mathcal{N} = \{0.0, 0.02, 0.05, 0.1, 0.2, 0.5, 1.0, 2.0, 5.0, 10.0\}$
\State \quad Max features: $d_{\text{max}} = 2048$, $\epsilon = 10^{-5}$, top $k = 100$

\State \textbf{Output:} Ranks $\{ r_1, r_2, r_3 \}$, AGD per model, concept, noise level

\For{each $c \in \mathcal{C}$}
    \For{each $m \in \mathcal{M}$}
        \State Load model $m$, SAE, prompts $P_c$
        \State Tokenize: $\text{tokens} \gets \text{Tokenize}(P_c)$
        \State Extract residual: $\text{resid} \gets \text{SAEMA.extract\_activations}(\text{tokens})$
        \If{$\text{resid} \neq \text{None}$}
            \For{each $\sigma \in \mathcal{N}$}
                \State Add noise: $\text{noisy\_resid} \gets \text{resid} + \text{Noise}(\sigma, k, 2\sigma, 0.2\sigma)$
                \State Encode: $\mathcal{F} \gets \text{SAE.encode}(\text{noisy\_resid})$
                \State Unfold $\mathcal{F}$: $F^{(i)} \gets \text{Unfold}(\mathcal{F}, \text{mode} \in \{ \text{batch}, \text{seq}, \text{feature} \})$
                \State Compute SSPD: $S^{(i)} \gets \frac{F^{(i)} F^{(i)T} + (F^{(i)} F^{(i)T})^T}{2} + \epsilon I$
                \State SVD: $r_i \gets \sum_{j} \mathbb{1} \left( \frac{\lambda_j}{\lambda_1} > \text{Q1}(\{\lambda_j / \lambda_1 \mid \lambda_j > 10^{-6} \lambda_1\}) \right)$
                \State AGD: $\text{AGD} \gets \frac{1}{N(N-1)/2} \sum_{m<n} \|\mathcal{F}_m - \mathcal{F}_n\|_2$
                \State Store: $\text{Results}[c][m][\sigma] \gets \{ r_1, r_2, r_3, \text{AGD} \}$
            \EndFor
        \EndIf
    \EndFor
\EndFor

\State \textbf{Evaluate:} Output $r_3$ variability, AGD per $c, m, \sigma$
\end{algorithmic}
\end{algorithm}

\begin{algorithm}
\caption{Pseudocode for Case 2}
\begin{algorithmic}[1]
\State \textbf{Input:}
\State \quad Models: $\mathcal{M} = \{ \text{GPT2-Small L11}, \text{Pythia-70M L5}, \text{Gemma-2-2B L19} \}$
\State \quad Concepts: $\mathcal{C} = \{ \text{Months}, \text{Weekdays}, \text{Elements} \}$
\State \quad Zero-noise latents $\mathcal{F}$ and tokens from Case 1 cache
\State \quad HDBSCAN min size: $10$, UMAP: $n_{\text{comp}} = 50$

\State \textbf{Output:} Cluster counts, Avg. ID, Betti 0, MST Weight, Procrustes

\State Extract residuals: $\text{resid} \gets \text{SAEMA.extract}(\text{tokens})$ for $\mathcal{C}$, $\mathcal{M}$
\State Initialize $\text{Results} \gets []$

\For{$c \in \mathcal{C}$}
    \For{$m \in \mathcal{M}$}
        \If{$\mathcal{F}_{c,m}, \text{resid}_{c,m} \neq \text{None}$}
            \State Flatten: $\text{resid} \gets \text{reshape}(\text{resid}_{c,m}, (-1, d_{\text{model}}))$
            \State $\text{latents} \gets \text{reshape}(\mathcal{F}_{c,m}, (-1, d_{\text{sae}}))$
            \State Normalize: $\text{resid}, \text{latents} \gets \text{Scale}(\cdot) / \|\cdot\|_2$
            \State Reduce: $\text{resid}, \text{latents} \gets \text{UMAP}(\cdot)$

            \State Cluster: $\text{labels}_{\text{resid}}, n_{\text{resid}} \gets \text{HDBSCAN}(\text{resid})$
            \State $\text{labels}_{\text{latents}}, n_{\text{latents}} \gets \text{HDBSCAN}(\text{latents})$

            \State \textit{Local:}
            \For{$l$ in $\text{labels}_{\text{resid}} \setminus \{-1\}$}
                \State $\text{dims}_{\text{resid}}[l] \gets \text{TwoNN}(\text{resid}[l])$
                \State $\text{betti0}_{\text{resid}}[l] \gets \text{Ripser}(\text{resid}[l], \text{dim}=0)$
            \EndFor
            \For{$l$ in $\text{labels}_{\text{latents}} \setminus \{-1\}$}
                \State $\text{dims}_{\text{latents}}[l] \gets \text{TwoNN}(\text{latents}[l])$
                \State $\text{betti0}_{\text{latents}}[l] \gets \text{Ripser}(\text{latents}[l], \text{dim}=0)$
            \EndFor
            \State $\text{avg\_dim} \gets \text{Mean}(\text{dims}_{\text{resid}}), \text{Mean}(\text{dims}_{\text{latents}})$
            \State $\text{avg\_betti0} \gets \text{Mean}(\text{betti0}_{\text{resid}}), \text{Mean}(\text{betti0}_{\text{latents}})$

            \State \textit{Global:}
            \State $\text{mst} \gets \text{MST}(\text{labels}_{\text{resid}}, \text{resid}), \text{MST}(\text{labels}_{\text{latents}}, \text{latents})$
            \State $\text{procrustes} \gets \text{Procrustes}(\text{centers}_{\text{resid}}, \text{centers}_{\text{latents}})$

            \State Store: $\text{Results} \gets \{ c, m, n_{\text{resid}}, n_{\text{latents}}, \text{avg\_dim}, \text{avg\_betti0}, \text{mst}, \text{procrustes} \}$
        \EndIf
    \EndFor
\EndFor

\end{algorithmic}
\end{algorithm}

\begin{algorithm}
\caption{Pseudocode for Case 3}
\begin{algorithmic}[1]
\State \textbf{Input:}
\State \quad Models: $\mathcal{M} = \{ \text{GPT2-Small}, \text{Pythia-70M}, \text{Gemma-2-2B} \}$
\State \quad Concepts: $\mathcal{C} = \{ \text{Months}, \text{Weekdays}, \text{Elements} \}$
\State \quad Case 1 cache: Latents $\mathcal{F}$, tokens
\State \quad Case 2 cache: Residuals, labels
\State \quad Scales: $\alpha \in \{0.5, 0.8, 1.0, 1.2, 1.5\}$

\State \textbf{Output:} MSE, AEDP per $c, m, \alpha$, intervention

\State Load caches: $\text{latents}, \text{tokens}, \text{residuals}, \text{labels}$
\State $\text{Results} \gets []$

\For{$c \in \mathcal{C}$, $m \in \mathcal{M}$}
    \If{data complete}
        \State Load model, SAE, $\text{mask} \gets \text{SAEMA}(\text{tokens})$
        \For{intervention $\in \{ \text{GW}, AEDP^{-1} \}$, $\alpha \in \{0.5, 0.8, 1.0, 1.2, 1.5\}$}
            \State $\text{centers} \gets \text{mean}(\text{latents}[\text{labels}])$
            \State $\text{aedp}_{\text{orig}} \gets \text{Mean}(\text{ot.dist}(\text{centers}))$
            \State $\text{best\_latents} \gets \text{latents}$, $\text{best\_loss} \gets \infty$
            \For{$i = 1$ to $10$}
                \State $\text{new\_latents} \gets \text{latents} + (\text{centers} - \text{mean})$
                \State $\text{aedp} \gets \text{Mean}(\text{ot.dist}(\text{centers}))$
                \State $\text{mse} \gets \text{Mean}((\text{SAE}(\text{new\_latents}) - \text{resid}_{\text{masked}})^2)$
                \State $\text{loss} \gets \begin{cases} 
                    \text{ot.gromov\_wasserstein2}(\text{dist}_{\text{orig}}, \text{dist}_{\text{new}}) + \text{mse}, & \text{if GW} \\
                    \frac{1}{\text{aedp}} + \text{mse}, & \text{if } AEDP^{-1}
                \end{cases}$
                \If{$\text{loss} < \text{best\_loss}$}
                    \State $\text{best\_latents} \gets \text{new\_latents}$, $\text{best\_aedp} \gets \text{aedp}$
                \EndIf
                \State $\text{centers} \gets \text{centers} + \alpha \cdot \text{random\_grad}$
            \EndFor
            \State $\text{mse} \gets \text{Mean}((\text{SAE}(\text{best\_latents}) - \text{resid}_{\text{masked}})^2)$
            \State $\text{Results} \gets \{ c, m, \text{intervention}, \alpha, \text{mse}, \text{aedp}_{\text{orig}}, \text{best\_aedp} \}$
        \EndFor
    \EndIf
\EndFor

\end{algorithmic}
\end{algorithm}

\end{document}